\documentclass[acmsmall]{acmart}
\pdfoutput=1

\usepackage{mathrsfs}

\PassOptionsToPackage{dvipsnames,table}{xcolor}
\usepackage[dvipsnames]{xcolor}
\usepackage{booktabs}
\usepackage{indentfirst}
\usepackage{amsmath}
\usepackage[linesnumbered,algoruled,boxed,lined]{algorithm2e}
\usepackage{tabularx}
\usepackage{array}
\usepackage{xspace}
\usepackage{breqn}
\usepackage[draft=true]{minted}
\usepackage{enumitem}
\usepackage{wrapfig}

\PassOptionsToPackage{hyphens}{url}
\usepackage{longtable}
\usepackage{calc}

\definecolor{lightgray}{HTML}{BDBDBD}
\definecolor{forestgreen}{HTML}{228B22}

\newcommand{\pw}{\texttt{prediction window}\xspace}

\newcommand{\lead}{\texttt{lead}\xspace}
\newcommand{\ie}{i.e.\ }

\newcommand{\aml}{AutoML}
\newcommand{\ptask}{prediction task}
\newcommand{\ds}{\textit{data scientist}\xspace}

\definecolor{mymaroon}{rgb}{0.5, 0.0, 0.0}
\definecolor{myOliveGreen}{rgb}{0.0, 0.4, 0.0}

\newminted[petel]{text}{formatcom=\color{mymaroon},autogobble=true,frame=none,fontsize=\footnotesize}

\makeatletter

\newcommand*{\tabminted@finalstrut}[1]{%
  \ifdim\prevdepth>0pt
    \ifdim\dp#1>\prevdepth
      \vskip\dimexpr(\dp#1)-\prevdepth\relax
    \fi
  \else
    \vskip\dimexpr(\dp#1)\relax
  \fi
}
\newcommand*{\@tabmintedend}{%
  \let\@finalstrut\tabminted@finalstrut
}
\makeatother

\newtheorem{defn}{Definition}[section]

\newenvironment{urlist}
{\begin{enumerate}[font=\footnotesize,before*=\footnotesize]

\setcounter{enumi}{0}
  \setlength{\itemsep}{1pt}
  \setlength{\parskip}{0pt}
  \setlength{\parsep}{0pt}
}
{\end{enumerate}}
\newcommand{\uref}[1]{[U\ref{#1}]}

\setcopyright{acmcopyright}
\copyrightyear{2018}
\acmYear{2018}
\acmDOI{10.1145/1122445.1122456}

\acmVolume{*}
\acmNumber{}
\acmArticle{111}
\acmMonth{8}

\begin{document}

\title{AutoML to Date and Beyond: Challenges and Opportunities}

\author{Shubhra Kanti Karmaker (``Santu'')}
\email{SKS0086@auburn.edu}
\affiliation{%
  \institution{Auburn University (Previously MIT LIDS)}
  \city{Cambridge}
  \state{MA}
}

\author{Md. Mahadi Hassan}
\email{sibathasan@gmail.com}
\affiliation{%
  \institution{Auburn University}
  \city{Auburn}
  \state{AL}
}

\author{Micah J. Smith}
\email{micahs@mit.edu}
\affiliation{%
  \institution{MIT LIDS}
  \city{Cambridge}
  \state{MA}
}

\author{Lei Xu}
\email{leix@mit.edu}
\affiliation{%
  \institution{MIT LIDS}
  \city{Cambridge}
  \state{MA}
}

\author{ChengXiang Zhai}
\email{czhai@illinois.edu}
\affiliation{%
  \institution{University of Illinois Urbana Champaign}
  \city{Urbana}
  \state{Illinois}
}

\author{Kalyan Veeramachaneni}
\email{kalyanv@mit.edu}
\affiliation{%
  \institution{MIT LIDS}
  \city{Cambridge}
  \state{MA}
}

\renewcommand{\shortauthors}{Shubhra Kanti Karmaker Santu, et al.}
\renewcommand{\shorttitle}{A Level-wise Taxonomic Perspective on AutoML}

\begin{abstract}
As big data becomes ubiquitous across domains, and more and more stakeholders aspire to make the most of their data, demand for machine learning tools has spurred researchers to explore the possibilities of automated machine learning (AutoML). AutoML tools aim to make machine learning accessible for non-machine learning experts (domain experts), to improve the efficiency of machine learning, and to accelerate machine learning research. But although automation and efficiency are among AutoML’s main selling points, the process still requires human involvement at a number of vital steps, including understanding the attributes of domain-specific data, defining prediction problems, creating a suitable training data set, and selecting a promising machine learning technique. These steps often require a prolonged back-and-forth that makes this process inefficient for domain experts and data scientists alike, and keeps so-called AutoML systems from being truly automatic. In this review article, we introduce a new classification system for AutoML systems, using a seven-tiered schematic to distinguish these systems based on their level of autonomy. We begin by describing what an end-to-end machine learning pipeline actually looks like, and which subtasks of the machine learning pipeline have been automated so far. We highlight those subtasks which are still done manually — generally by a data scientist — and explain how this limits domain experts’ access to machine learning. Next, we introduce our novel level-based taxonomy for AutoML systems and define each level according to the scope of automation support provided. Finally, we lay out a roadmap for the future, pinpointing the research required to further automate the end-to-end machine learning pipeline and discussing important challenges that stand in the way of this ambitious goal.

\end{abstract}

\maketitle

\section{Introduction}\label{sec:intro}
One of the main consequences of the massive increase in web-based systems has been the proliferation of “big data” — automatically generated information that can shed light on user demographics, behaviors, and needs. Companies across domains and applications are increasingly focused on exploiting this data to improve their products and services. The rapid growth of machine learning infrastructure and high-performance computing have significantly contributed to this process. Business entities are hiring more and more data scientists (5X growth) and ML engineers (12X growth) in order to better make sense of their data \cite{linkedin}, while both industry and academia have made \emph{artificial intelligence} (AI) and \emph{machine learning} (ML) research priorities.

As these stakeholders attempt to make the most of their data, demand for machine learning tools has spurred researchers to explore the possibilities of automated machine learning (\aml). \aml is essentially a paradigm for automating the application of machine learning to real-world problems (what we call the "end-to-end ML process"). Although automation and efficiency are among \aml{'s} main selling points, this process still requires a surprising level of human involvement. To better understand the problem, let us discuss what an end-to-end ML process looks like today. 

Generally, a business or enterprise adopts machine learning in order to fulfill a specific need. For instance, say an organization's leadership team launches a product or service, and customers start interacting with it. Domain experts begin digging deeper into customer behavior, and may deploy sensors and conduct experiments to closely monitor different components of the product. Because all of these transactions, customer interactions, experiments and sensors produce a huge amount of data every day — too much for these domain experts to manually obtain insights from — data scientists are brought in to help. As domain experts and data scientists are the key users of \aml tools, we pause here to define their roles: 

\begin{itemize}[leftmargin=*]
    \item {\bf Domain Expert:} A person who is fluent in the domain where ML is being applied, but has minimal knowledge of how ML itself works.
    \item {\bf Data Scientist:} A person who knows how ML works, but has minimal knowledge of the domain where it is being applied.
\end{itemize}

The data scientist must work with the domain expert in order to understand the context in which the data was collected, as well as the overall goals. S/he then translates the goal into a computational \ptask, extracts useful features from the raw data, and selects a relevant ML model to solve the \ptask. A number of vital steps of this ML process -- including understanding the attributes of domain-specific data, defining prediction problems, and creating a suitable training data set -- are generally still done manually by the data scientist on an ad-hoc basis. This often requires a lot of back-and-forth between the data scientist(s) and domain expert(s), making the whole process more difficult and inefficient. 

\begin{figure}
  \centering
    \caption{ A typical end-to-end machine learning process, with multiple stakeholders bringing their individual expertise, tools, methods, datasets and goals. }
    \label{fig:pipeline}
    \includegraphics[width=0.65\textwidth]{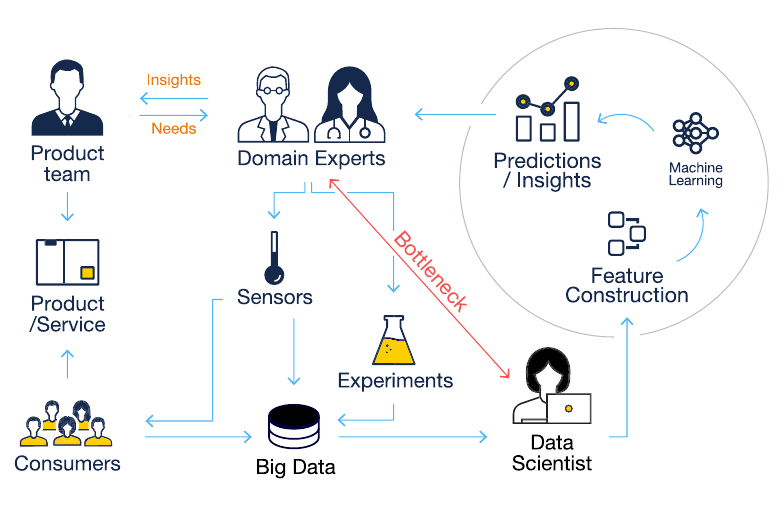}
\vspace{-5mm}
\end{figure}

It is only once all of these steps are complete that existing \aml{} systems come in -- automatically applying and tuning machine learning techniques to the training / testing datasets that have been compiled by the human data scientist, in order to solve a prediction task that has been carefully defined by the same person. The data scientist is then called upon again, this time to draw valuable insights from the data and models. S/he then reports these results to the domain experts and business leaders, who use them to improve the product and address any issues with it. This whole process is laid out visually in Figure~\ref{fig:pipeline}.

The term \aml is commonly understood to denote an system that can perform the common data science tasks described above with minimal human intervention. However, current \aml{} systems with that designation are actually far from achieving that level of automation.  We posit that the ``northstar'' goals for \aml{} systems should be twofold: to increase the efficiency of data scientists, and to enable domain experts to directly use machine learning using automation tools, inserting their domain knowledge as necessary. The latter goal, if achieved, will bring along a welcome secondary effect -- making the powerful tool of machine learning more accessible to non-experts across fields.

\noindent \textbf{How do we achieve this ``\textit{northstar}'' goal}: To achieve this ``\textit{northstar}'', we must expand the definition of ''\aml'' systems, in order to speak more precisely about these tools and plan for future improvements. In this review article, we present a new level-based classification system for current and future AutoML solutions. We first describe the tasks a data scientist performs during the end-to-end machine learning process (see Section~\ref{sec:pipeline}), pointing out subtasks which are still generally done manually as well as those that current \aml{} systems can perform automatically. We highlight the overall flow, iterative loops within this process, and the sources of bottlenecks (see Section~\ref{flow}). We then present our level-based classification system, which is based on the automation of particular subtasks, and provide examples of existing \aml{} systems at each level (see Section~\ref{levels}). Finally, we lay out a road map for the future, pinpointing the research required to further automate the end-to-end machine learning process and discussing important challenges that stand in the way of this ambitious goal (see Section~\ref{future}).

\vspace{-2mm}
\section{What does a Data Scientist Do?}\label{sec:pipeline}
As shown in the representative end-to-end machine learning process in Figure~\ref{fig:pipeline}, \emph{data scientists} often perform a well-defined set of functions in order to achieve analysis goals. We formalize these steps below. Over the past decade, the machine learning and systems communities have worked to automate many, but not all, of these steps.


{
\begin{itemize}[leftmargin=*]

    \item {\bf Task Formulation (TF):} In any \textit{data science} endeavor, the \ds first interacts with the domain expert to understand the problem at hand, examines the available data, and formulates a machine learning task to help solve this problem. Formulating the task can involve a lot of back and forth, where the \ds and domain expert must consider multiple possibilities and check if the required data is available each time before making a decision. This is currently a manual process, and is often long and arduous, impeded by mismatched expectations and communication gaps.



\item {\bf Data Visualization, Cleaning and Curation (DCC):} Throughout the analysis process, data scientists repeatedly return to the original data in order to propose and revise models and test hypotheses. This often requires cleaning and curating new subsets of data. After identifying the relevant data, data scientists may handle missing values,  ``\textit{join}'' multiple datasets and perform other functions for improving data quality. They may also create visualizations to direct this process and aid in discussions. The \textit{data mining}, \textit{systems} and \textit{database} communities have progressed appreciably in automating this step, efforts that are well summarized in~\citet{ilyas2015trends} and~\citet{chu2016data}. In several of these approaches, machine learning itself is used to automatically perform tasks involved in curation, cleaning and dataset linking. More recent approaches have focused on data cleaning challenges for specific domains, including electronic health records (\citet{miao2018assessment}), time series data (\citet{zhang2017time}), online reviews (\citet{minnich2016clearview}), and wireless sensor networks (\citet{cheng2018data}). Another school of effort has primarily focused on interactive or active data cleaning (e.g.,~\citet{krishnan2016activeclean,chu2015katara,haas2015wisteria}). For a more detailed review of existing literature on data visualization, cleaning and curation, refer to Appendix~\ref{appen:DCC}.

\item {\bf Prediction Engineering (PE):} Prediction engineering involves constructing and assigning labels to data points according to the set \ptask, and creating meaningful training and testing sets as collections of <\texttt{data\_point}, \texttt{label}> tuples. This task is usually done manually by data scientists, sometimes with the help of human data annotators. However, there is a growing demand for automated \textit{prediction engineering}. Kanter et.al.~\cite{kanter2016label} tried to automate this process (the first attempt of this kind) by proposing a generalized three-part framework — Label, Segment, Featurize (L-S-F) — which provides abstractions that enable data scientists to customize the \emph{prediction engineering} process for unique prediction problems.

\item {\bf Feature Engineering (FE):} By the end of the first three steps, a data scientist will have defined a ML task (most commonly a predictive task), created training and testing sets, and begun the model building process — the first step of which is to attempt to construct informative features from raw data. This process is called ``feature engineering,’’ which we define formally as follows. 

\begin{defn}
\textit{Feature engineering} is the process of transforming raw data into features that can better represent an underlying problem for computational predictive models, resulting in improved model accuracy on unseen data.
\end{defn}

Later, these features can be directly fed to ML models in order to train them and make predictions. In the past 5 years, a number of efforts have focused on automating the process of \textit{feature engineering} (\citet{katz2016explorekit,kanter2015deep,mountantonakis2017linked,van2017automatic,khurana2017feature,kaul2017autolearn}). There are now multiple systems and open source libraries that focus on extracting useful features from data and generating feature matrices to construct the training set. For more details on these efforts to automate \textit{feature engineering}, refer to Appendix~\ref{appen:FE}.

\item {\bf Machine Learning (ML):} Once feature engineering is complete, the data scientist selects a suitable machine learning technique based on the \ptask, and begins training models. These models consist of basic machine learning techniques -- such as decision trees, support vectors machines, linear regressions, neural networks, etc. -- which have current implementations across various software packages including scikit-learn~\cite{scikit-learn} and weka~\cite{witten2002data}. One crucial task associated with training is optimizing, or \textit{tuning}, any hyperparameters associated with the learning model.

\begin{itemize}
    \item[--]{\bf Hyperparameter Tuning:} Machine learning models often contain multiple hyperparameters, the values of which are critical for obtaining good performance. Data scientists must tune these hyperparameters in order to find models that work reasonably well. Efforts to automate tuning include \citet{bergstra2012random,snoek2012practical,hutter2011sequential,bergstra2011algorithms,bengio2012practical,bergstra2013making,swersky2013multi,maclaurin2015gradient}. For a deeper discussion of \emph{hyperparameter tuning} literature, refer to Appendix~\ref{appen:HT}.
\end{itemize}

\item {\bf Alternative Models Exploration, Testing and Validation (ATV): } 
A \ptask can generally be solved using multiple ML techniques. Currently, automation work related to the selection, validation and finalization of models is broadly categorized under the subfield \textit{AutoML}. We are instead referring to this step as Alternative Models Exploration, Testing and Validation (ATV), as we view \textit{AutoML} to be an intelligent solution with the far broader goal of automating an end-to-end machine learning process. Perhaps the past decade’s most widely studied area, \footnote{A quick search of “AutoML” in Google Scholar will bring up thousands of papers, with the top results cited over 100 times.} major efforts in this direction include~\citet{thornton2013auto,feurer2015efficient,swearingen2017atm}. A number of commercial systems already exist for this task. Below we briefly describe the two subtasks of ATV, and point the reader to some automation systems, open source libraries and commercial systems.

\begin{itemize}

\item[--] {\bf Alternative Models Exploration:} Data scientists often spend time exploring alternative models and deducing their pros and cons. A recent DARPA-led initiative, ``Data-Driven Discovery of Models'' (D3M), focuses on alternative strategies for model exploration, or the automation of ``complete'' data mining (\citet{lippmann2016d3m}). It has inspired several other tools, such as AlphaD3M, an \aml{} system based on edit operations performed over machine learning pipeline primitives (\citet{drori2018alphad3m}), and ML Bazaar, a composable framework for developing what are called ``end-to-end'' ML systems (\citet{smith2019mlbazaar}). In a similar vein, \citet{olson2016evaluation} developed a tree-based pipeline optimization tool, TPOT, that automatically designs and optimizes machine learning pipelines for a given domain, while \citet{cashman2018visual} studied how visual analytics can play a valuable role in the automated model discovery process. Following the growing interest in deep neural networks, multiple researchers including \citet{zoph2016neural,zoph2017learning,liu2017hierarchical,liu2017progressive,real2018regularized,pham2018efficient,baker2017accelerating} have proposed neural or reinforcement learning models to automatically search optimal neural network structures. For a more detailed review of existing literature on alternative models exploration, refer to Appendix~\ref{appen:AME}.

\item{\bf Evaluation:} After selecting the best-performing model for the training set, the data scientist must evaluate this model on the testing set and report the statistical significance of these evaluations. Because this step is so important, the machine learning community has put significant effort into comparative evaluation of different machine learning models. Some recent efforts include benchmarking of deep learning software tools by \citet{shi2016benchmarking}, hyperparameter auto-tuning methods by \citet{Laura:2018}, deep reinforcement learning approaches by \citet{duan2016benchmarking} and bio-medical data mining methods by \citet{olson2017pmlb}. More details about evaluation and benchmarking can be found in Appendix~\ref{appen:E}.

\end{itemize}

\item {\bf Result Summary and Recommendation (RSR):} The final and arguably most important job of a data scientist is to summarize the findings and recommend the most useful tasks to domain experts. For each task, recommendations can be made at the level of models, features, or computational overhead. This step is unsystematic and done mostly manually.

\end{itemize}
}

The different libraries and tools discussed so far typically make up one component within a larger system that aims to manage several practical aspects of machine learning, such as parallel and distributed training, tuning, model storage, and even serving and deployment. Such large systems include ATM \citet{swearingen2017atm}, Vizier \citet{golovin2017google}, and Rafiki \citet{wang2018rafiki}, as well as commercial systems like Google AutoML\footnote{https://cloud.google.com/automl/} and Amazon Forecast\footnote{https://aws.amazon.com/blogs/aws/amazon-forecast-time-series-forecasting-made-easy/}. For more details about these complex machine learning systems, refer to Appendix~\ref{appen:system}.

Table~\ref{tab:automl} provides a comprehensive summary of current tools that aim to automate particular subtasks (or sets of subtasks) of the machine learning process, with more details in the appendix. It is organized by subtask, and lists the notable research contributions, survey articles, open source systems and commercial systems that relate to that subtask.\\

{\footnotesize
\newlength{\tabfivecol}
\setlength{\tabfivecol}{\linewidth - \tabcolsep * 10}

\setlength\LTleft{0pt}
\setlength\LTright{0pt}
\begin{longtable}{@{} @{\extracolsep{\fill}} >{\raggedright\arraybackslash}p{0.15\tabfivecol} >{\raggedright\arraybackslash}p{0.21\tabfivecol} >{\raggedright\arraybackslash}p{0.21\tabfivecol} >{\raggedright\arraybackslash}p{0.23\tabfivecol} >{\raggedright\arraybackslash}p{0.21\tabfivecol} @{}}
\caption{\label{tab:automl} A summary of contributions in the broader field of automated machine learning (AutoML). For each subtask, we list related and notable research contributions, survey articles, open source systems and commercial systems. For more details, refer to the Appendix.}\\
\toprule
Task & Research Contributions & Survey Articles & Open Source Systems & Commercial Systems\\
\midrule
\endfirsthead
\multicolumn{5}{c}{\tablename\ \thetable\ -- \textit{Continued from previous page}} \\
\toprule
Task & Research Contributions & Survey Articles & Open Source Systems & Commercial Systems\\
\midrule
\endhead
\bottomrule
\endfoot
Data Visualization,  & \citet{miao2018assessment} & \citet{ilyas2015trends} & Open Refine \uref{OpenRefine}  & Trifacta Wrangler \uref{TrifactaWrangler}\\
Cleaning and Curation & \citet{zhang2017time} & \citet{chu2016data} & Drake \uref{Drake} & TIBCO Clarity \uref{TIBCOClarity}\\
 & \citet{minnich2016clearview} &  & Open Source Data Quality and Profiling \uref{DataQuality} & Winpure \uref{Winpure}\\
 & \citet{cheng2018data} &  & Talented Data Quality  \uref{TalentedDataQuality} & Data Ladder \uref{DataLadder}\\
 & \citet{krishnan2016activeclean} &  &  & Data Cleaner \uref{DataCleaner}\\
 & \citet{chu2015katara} &  &  & Cloudingo \uref{Cloudingo}\\
 & \citet{haas2015wisteria} &  &  & Refier \uref{Refier}\\
 &  &  &  & IBM Infosphere Quality Stage \uref{IBMInfosphereQualityStage}\\
\midrule
Feature & \citet{katz2016explorekit} & \citet{wang2005survey} & FeatureTools \uref{FeatureTools} & FeatureTools \uref{FeatureTools}\\
Engineering & \citet{kanter2015deep} & \citet{chandrashekar2014survey} & featexp \uref{featexp} & \\
 & \citet{mountantonakis2017linked} & \citet{ji2011survey} & MLFeatureSelection \uref{MLFeatureSelection} & \\
 & \citet{van2017automatic} & \citet{sheikhpour2017survey} & Feature Engineering \& Feature Selection \uref{FeatureEngineeringFeatureSelection} & \\
 & \citet{khurana2017feature} & \citet{liu2015feature} & FeatureHub \uref{FeatureHub} & \\
 & \citet{kaul2017autolearn} &  & featuretoolsR \uref{featuretoolsR} & \\
 &  &  & Feast \uref{Feast} & \\
 &  &  & ExploreKit \uref{ExploreKit} & \\
\midrule
Hyper\-parameter  & \citet{bergstra2012random} & \citet{vanschoren2018meta} & Hyperparameter Hunter \uref{HyperparameterHunter} & Amazon Sagemaker \uref{AmazonSagemaker}\\
Tuning & \citet{snoek2012practical} & \citet{hutter2015beyond} & hyperband \uref{hyperband} & BigML OptiML \uref{BigMLOptiML}\\
 & \citet{hutter2011sequential} &  & Hyperboard \uref{Hyperboard} & Google HyperTune \uref{GoogleHyperTune}\\
 & \citet{bergstra2011algorithms} &  & SHERPA \uref{SHERPA} & Indie Solver \uref{IndieSolver}\\
 & \citet{bengio2012practical} &  & Milano \uref{Milano} & Mind Foundry OPTaaS \uref{MindFoundryOPTaaS}\\
 & \citet{bergstra2013making} &  & BBopt \uref{BBopt} & sigopt \uref{sigopt}\\
 & \citet{swersky2013multi} &  & adatune \uref{adatune} & \\
 & \citet{maclaurin2015gradient} &  & gentun \uref{gentun} & \\
 &  &  & optuna \uref{optuna} & \\
 &  &  & test-tube \uref{test-tube} & \\
 &  &  & Advisor \uref{Advisor} & \\
\midrule

Alternative Models  & \citet{thornton2013auto} & \citet{he2019automl} & AutoKeras \uref{AutoKeras} & H2O Driverless AI \uref{H2ODriverlessAI}\\
Exploration & \citet{feurer2015efficient} & \citet{elsken2018neural} & AdaNet \uref{AdaNet} & Cloud AutoML \uref{CloudAutoML}\\
 & \citet{swearingen2017atm} & \citet{zoller2019survey} & PocketFlow \uref{PocketFlow} & C3 AI Suite \uref{C3AISuite}\\
 & \citet{lippmann2016d3m} &  & automl-gs \uref{automl-gs} & FireFly \uref{FireFly}\\
 & \citet{zoph2016neural} &  & MLBox \uref{MLBox} & Builton \uref{Builton}\\
 & \citet{zoph2017learning} &  & Morph-net \uref{Morph-net} & Determined AI Platform \uref{DeterminedAIPlatform}\\
 & \citet{liu2017hierarchical} &  & ATM \uref{ATM} & \\
 & \citet{liu2017progressive} &  & TransmogrifAI \uref{TransmogrifAI} & \\
 & \citet{real2018regularized} &  & RemixAutoML \uref{RemixAutoML} & \\
 & \citet{pham2018efficient} &  &  & \\
 & \citet{baker2017accelerating} &  &  & \\
\midrule
Evaluation and  & \citet{shi2016benchmarking} & \citet{gijsbers2019open} & OpenML100 \uref{OpenML100} & \\
Benchmarking & \citet{Laura:2018} & \citet{klein2019meta} & OpenML-CC18 \uref{OpenML-CC18} & \\
 & \citet{duan2016benchmarking} &  & AutoML Challenges \uref{AutoMLChallenges} & \\
 & \citet{olson2017pmlb} &  &  & \\
 & \citet{eggensperger2015efficient} &  &  & \\
 
 \midrule
 Systems & \citet{swearingen2017atm} &  & Scikit-learn & DotData \uref{DotData}\\
 & \citet{golovin2017google} &  & TPOT \uref{TPOT} \cite{olson2016evaluation} & IBM AutoAI \uref{IBMAutoAI}\\
 & \citet{wang2018rafiki} &  & Auto-Sklearn \uref{Auto-Sklearn} & Azure Machine Learning \uref{AzureMachineLearning}\\
 &  &  & AlphaD3M \cite{drori2018alphad3m} & Google Could AI platform \uref{GoogleCouldAIplatform}\\
 &  &  & ML Bazaar \cite{smith2019mlbazaar} & Amazoon AWS service \uref{AmazoonAWSservice}\\
 &  &  & PyTorch & Data Robot \uref{DataRobot}\\
 &  &  &  & Amazon Forecast \uref{AmazonForecast}\\
 &  &  &  & RapidMiner \cite{hofmann2016rapidminer}\\
  &  &  &  & Orange \cite{demvsar2012orange}\\
\end{longtable}

}

\section{The Current Flow of End-to-End Machine Learning Pipelines}\label{flow}
 
\begin{figure*}
  \centering
    \caption{A flowchart showing the machine learning process. This chart highlights points of interaction between domain experts and data scientists, along with bottlenecks.}
    \label{fig:cycle}
    
    \includegraphics[width=0.8\textwidth]{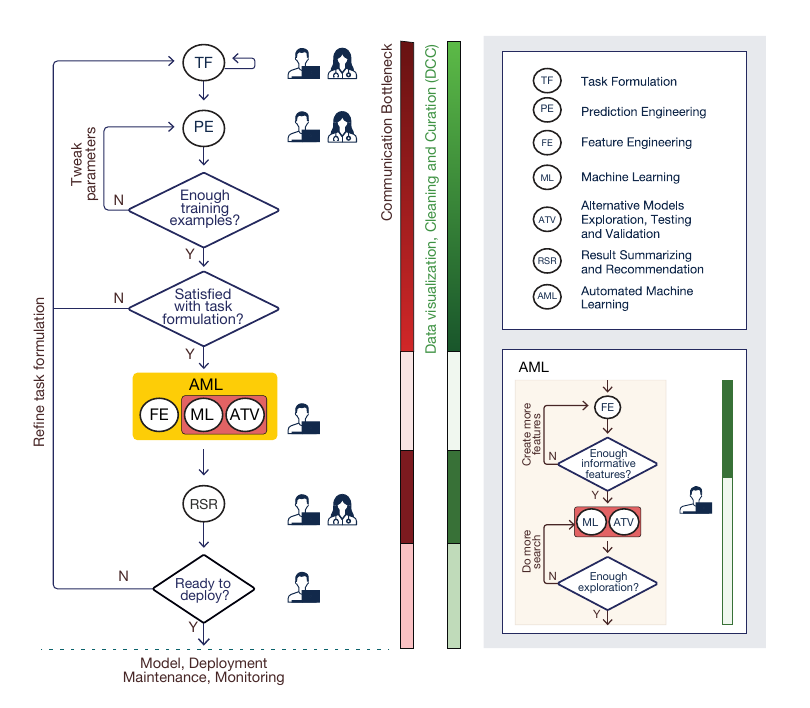}
\vspace{-5mm}
\end{figure*}

The primary goal of \textit{AutoML} is to reduce the manual effort involved with machine learning technologies, thus accelerating their deployment. Consequently, various systems have attempted to minimize the work required to perform certain steps of the machine learning development workflow (see table~\ref{tab:automl} and the appendix for a detailed discussion). For example, DeepDive / Snorkel~\cite{ratner2020snorkel} is a general, high-level workflow support system that helps users label and manage training data and provides high-level support for model selection. As previously mentioned, however, developing ML solutions still involves a lot of manual work. To design a truly automated system, it is important to address the bottlenecks in the current process. To better visualize these bottlenecks, we present in Figure~\ref{fig:cycle} a flowchart showing the end-to-end machine learning process. For each step in the flow, we outline the role of domain experts, the amount of manual work performed by the data scientist, and the communication required between the two.

Figure~\ref{fig:cycle} highlights key subtasks that need more attention from the research community if full automation is to be achieved. Models and software developed so far have enabled or made significant progress towards the automation of data visualization, cleaning and curation (DCC), machine learning (ML), feature engineering (FE) and alternative models exploration, testing and validation (ATV). The biggest communication bottlenecks currently happen in two places: Task formulation (TF) and result summary and recommendation (RSR), both of which require a significant amount of manual work. Prediction engineering (PE) also can lead to prolonged back-and-forth between the data scientist and the domain expert. 

These tasks are still largely unstructured and are performed manually through trial and error. The community has been reluctant to focus on them, possibly due to their human-centric nature -- these tasks involve significant human interaction, and their evaluation is also largely subjective. For example, without a robust user preference model, it is hard to identify ``interesting'' prediction tasks and eventually make useful automatic recommendations. A lack of systematic studies of these problems prevent the completion of a fully automated pipeline that can transform a dataset into a useful predictive model and use it to mine insights from it. 

It should be noted that existing AutoML solutions have primarily approached these tasks from a software systems perspective without much theoretical analysis. Another important thing to note is that domain experts currently have minimal access to the FE, ML and ATV sub-tasks, and as a result are left out of these parts of the process — instead waiting for the data scientist to figure it out and get back to them, perhaps with some exciting results. Finally, once business leaders agree to deploy a particular predictive model, deployment, maintenance and monitoring are also handled by data scientists and software engineers.

\vspace{-2mm} 
\section{Classifying Automl systems based on levels of automation}\label{levels}\vspace{-1mm} 
\begin{figure}[!t]
  \centering
    \caption{Levels of automation possible for end-to-end machine learning endeavors. Level 0, the lowest level, is entirely manual and requires people to develop and write software for the whole process. Level 6, the highest level, is entirely automated from task formulation all the way to result summary and recommendation. This figure provides examples of existing systems at each level, along with a color gradient indicating accessibility to the domain expert. As is shown here, increased automation has two simultaneous effects. First, it enables domain experts to make use of machine learning. Second, it increases the productivity of data scientists, because their manual work declines as the levels increase.}
    \label{fig:Levels}
    \vspace{1mm}
    \includegraphics[width=0.86\textwidth]{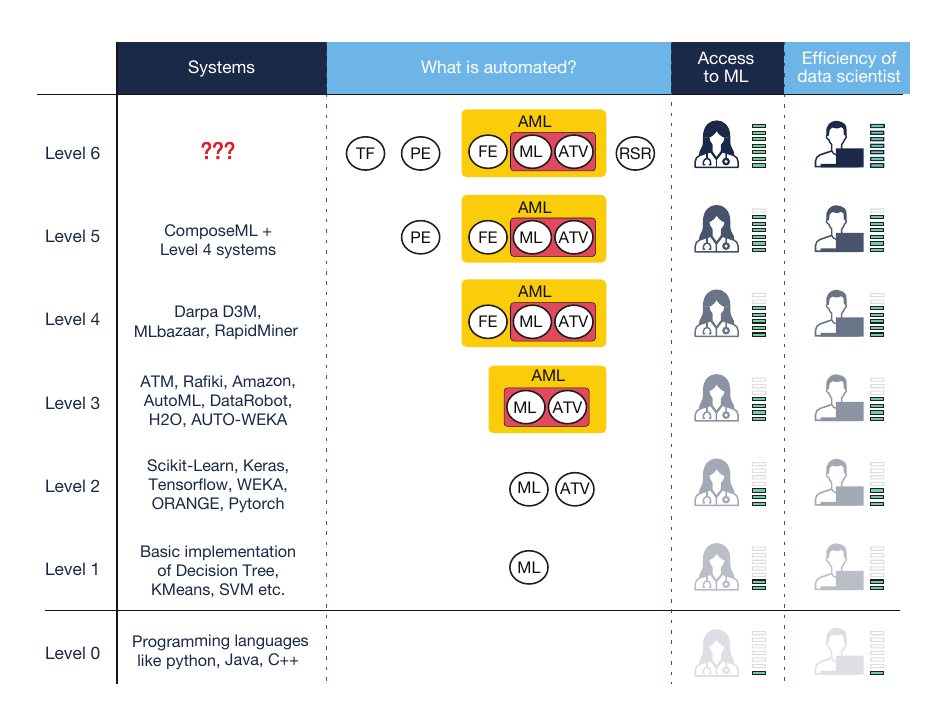}
    \vspace{-5mm}
\end{figure}

To better characterize the current drive toward automation by the ML community, we first define a set of taxonomic levels to represent different degrees of automation provided by systems developed so far. These levels are designed such that lower levels mean less automation and more manual work, and higher levels mean more automation and less manual work. A system’s level also correlates with a domain expert's ability to interact with that system despite a lack of ML expertise. Figure~\ref{fig:Levels} illustrates the proposed taxonomic levels, which we define below. These levels have no absolute meaning, but are rather a way to organize machine learning solutions based on the degree of automation they provide.

\vspace{-2mm} 
\subsection{ \textit{\textbf{Level 0 - No automation}}}\vspace{-2mm} 
All ML tasks are performed manually by hand-coded implementations. No automation is supported. 

\vspace{1mm}
{\textbf{Real-World Example:}} Machine learning researchers, who study and develop new machine learning models, mostly work at this level as they build core machine learning algorithms from scratch.

\vspace{-2mm}  
\subsection{\textit{\textbf{ Level 1 - Only ML automated }}}\vspace{-2mm}
These tools provide basic implementations of core machine learning algorithms. This is the minimal level of automation supported by a system. Tools at this level are generally developed by machine learning researchers and used by data scientists.

\vspace{1mm}
{\textbf{Real-World Example:}} C++ implementation of Support Vector Machine Classifier, which can be used as a library by data scientists. However, training set construction, label creation, feature construction, hyperparameter tuning, etc. must be carried out manually by the data scientist.  
    
\vspace{-2mm}
\subsection{\textit{\textbf{Level 2 - ML + ATV automated separately}}}\vspace{-2mm}
While these automated tools are popular with data scientists, they require a lot of manual work and are not at all accessible to domain experts.

\vspace{1mm}
{\textbf{Real-World Example:}} When a training set and well-defined feature sets are already available, data scientists can use a Level 2 tool like Weka to perform the actual learning and hyperparameter tuning tasks.

\vspace{-2mm} 
\subsection{\textit{\textbf{ Level 3 - ML + ATV automated jointly}}}\vspace{-2mm}
Some automated tools are provided at this level, while the rest of the work is done manually by the data scientist. Domain experts still do not have direct access to tools at this level.

\vspace{1mm}
{\textbf{Real-World Example:}} Level 3 tools like AutoWeka allow users to perform learning and hyperparameter tuning jointly, and thus require less manual work than Level 2 tools.

\vspace{-2mm} 
\subsection{\textit{\textbf{Level 4 - ML + ATV + FE automated}}}\vspace{-2mm} 
A domain expert can interact minimally with a system at this level. However, the data scientist must still perform a significant amount of manual work.

\vspace{1mm}
{\textbf{Real-World Example:}}  Consider a  wind energy company which owns a number of turbines, each equipped with many sensors. Predicting turbine failure in advance can save the company headaches and reduce their costs. However, numbers generated by the sensors are only meaningful to wind energy experts. A data scientist working on this problem must spend long hours with these experts, in order to understand the sensor data and what indicates turbine failure so that s/he can create meaningful features. Level 4 tools like MLBazaar aid the data scientist by automating this feature construction step. They can also actively engage domain experts by asking them to verify the quality of automatically created features.


\vspace{-2mm} 
\subsection{\textit{\textbf{ Level 5 - ML + ATV + FE + PE automated }}}\vspace{-1mm} 
Domain experts can comfortably interact with systems at this level, and minimal manual work is required from the data scientist.
    
\vspace{1mm}
{\textbf{Real-World Open Problem and Potential Application:}}  Imagine a railway company that continuously monitors signals from different engine sensors, and wants to predict future failure of engine components. A data scientist working on this problem must collaborate with domain experts, including engineers, programmers, and even conductors, to fully understand how failure is defined and create effective training examples for learning the predictive model. A Level 5 AutoML tool will be very useful for everyone in this scenario, because it can help users automatically create a good quality training set for supervised learning, as well as a validation set for testing purposes.

Automatic prediction engineering distinguishes a Level 5 AutoML solution from those at lower levels. There are many open problems associated with automatic \textit{prediction engineering} because it requires identifying missing information from natural language problem descriptions and filling them with of computable metrics. For example, consider the problem of forecasting future engine failures due to inclement weather. Two fundamental  challenges here involve: 1) Computationally defining ``inclement'' weather and 2) Automatically collecting the ``right'' training examples from historic data. Thus, automatic prediction engineering is essentially a hybrid human-computer interaction and data science problem, and will require joint investigation from both of these communities.


\vspace{-2mm} 
\subsection{\textit{\textbf{Level 6 - ML + ATV + FE + PE + TF + RSR automated}}}\vspace{-1mm} 

Level 6 provides the maximum amount of automation and requires the least amount of manual work. Domain experts can easily use Level 6 systems by themselves.
    
\vspace{1mm}
{\textbf{Real-World Open Problem and Potential Application:}}  Imagine a hospital that collects patient electroencephalography (EEG) results on a regular basis, and wants to use them to predict which patients are at higher risk for epilepsy. Doctors are not experts at building such predictive models; however, they do have enough medical knowledge to understand EEG log data and define prediction tasks rigorously. Level 6 AutoML tools allow doctors and stakeholders to directly input their goals, and can formulate relevant prediction problems almost autonomously and recommend relevant prediction tasks to users along with the corresponding results.

The two key functionalities distinguishing Level 6 automation tools from those on lower levels are: 1) \textit{Automatic Prediction Task Formulation} and 2) \textit{Recommendation}. Both functionalities come with numerous open questions that warrant rigorous academic investigation. For example, one way to formulate \textit{automatic prediction task formulation} is to frame it as an information extraction task within a dialog-based framework. In this case, challenges include but are not limited to:  How can we develop a general representation for a machine learning task? How can we extract task-specific parameters through active dialog with users? How can we efficiently navigate a very high-dimensional search space containing all possible prediction tasks?

Although recommendation is more of a classic \textit{information retrieval} problem, the prospect of automating it invites the following questions: How can we evaluate the quality of a prediction task itself? How can we revise a recommendation model by leveraging active user feedback? How can we capture business goals while recommending prediction tasks? And so on.

The AutoML ecosystem is sprawling. Achieving full automation will require expertise from many subfields. Unfortunately, most existing AutoML solutions focus almost entirely on hyperparameter tuning and feature engineering, and have primarily approached the problem from a software engineering and integration perspective. In developing our taxonomy, we aim to highlight core theoretical problems associated with different levels of machine learning automation, and invite people to recognize and investigate them systematically.


\vspace{-2mm} 
\subsection*{\underline{Using our tiered taxonomy to classify AutoML tools.}}\vspace{-1mm} 
Classifying a machine learning tool into one of these seven levels simply requires identifying which data science subtasks (as presented in section~\ref{sec:pipeline}) it has automated. The subtasks themselves act as features for classifying the AutoML tools, where each feature assumes a binary value: `yes' means automation is provided, while `no' means it is not. For example, MLbazaar~\cite{smith2019mlbazaar} provides automation for ML (machine learning), ATV (alternative models exploration, testing and validation) and FE (feature engineering), but does not provide automation for TF (task formulation), PE (prediction engineering) and RSR (result summary and recommendation). Thus, MLbazaar is categorized as an AutoML system with Level 4 automation. As another example, the popular data mining tool WEKA~\cite{witten2002data} provides automation for ML and ATV separately, and thus will be considered an AutoML system with Level 2 automation. An extension of WEKA, i.e., AUTO-WEKA~\cite{thornton2013auto}, provides automation for ML and ATV separately and thus will be categorized as a Level 3 tool.

\section {Towards Level 6 Autonomy for AutoML}\label{future}
\vspace{-2mm}

As Figure~\ref{fig:Levels} clearly demonstrates, automation efforts so far have primarily been limited to tools that belong in Levels 1 through 4. The field is sorely lacking in both Level 5 and Level 6 automation tools -- the only ones that allow domain experts to contribute heavily and take the load off of data scientists. 

How can we make the machine learning process more efficient and automatic, and relieve the communication bottleneck between data scientists and domain experts? One option is to enable domain experts to directly engage with big data without requiring them to have detailed knowledge of machine learning. Although this is a highly ambitious goal, if reached, it will expedite the implementation of machine learning systems and allow a far broader audience to make use of ML. Imagine an intelligent agent that can guide domain experts in big data exploration and analytical tasks, taking care of certain subtasks automatically and thus filling the role usually played by a \emph{human} data scientist. This kind of system would make big data analytics more appealing and accessible to business entities, and increase the productivity of human data scientists significantly by automating a large portion of their manual labor. Only a Level 6 AutoML tool could provide such support.

The importance of \textit{task formulation} (TF) -- a subtask that is not currently automated -- and \textit{prediction engineering} (PE) -- one that rarely is -- have been substantiated by leading practitioners in the software industry. In popular tutorial materials, practitioners discuss the value of TF and PE, and their high impact on the use and value of the overall ML model \cite{facebookfieldguide}. For example, someone seeking to predict user enjoyment of a mobile app -- which is difficult to define and model -- can get a similar answer by focusing on a simpler, easier-to-define metric like ``number of user sessions per week.'' According to these practitioners, defining the right prediction goal is often more important than choosing a particular algorithm, and a few hours spent at this stage in the process can save many weeks of work downstream.

Previous work has laid some foundations for automating \emph{prediction engineering}. \citet{schreck2016would} designed a formal language, called \emph{Trane}, to describe prediction problems over relational datasets, allowing data scientists to specify problems in that language. Another related work, MLFriend~\cite{xu2019mlfriend}, is an interactive framework that defines useful prediction problems on event-driven time-series data, and is designed to appeal to domain experts. In addition, Microsoft has introduced an alternative training paradigm, \emph{Machine Teaching}, which actively teaches domain experts to train machine learning models~\cite{simard2017machine} The authors have proposed a set of principles for teaching those new to the machine learning field, including a universal teaching language, ``realizability'' of all target concepts, a rich and diverse sampling set, and distribution robustness. The authors posit that bringing new people on board through active teaching will accelerate innovation and empower millions of new uses for machine learning models. While these are steps toward the goal of an end-to-end automated machine learning pipeline, they are limited by their focus on a particular data science function (e.g. Machine Teaching focuses on the training process, while MLFriend and Trane focus on defining the prediction task) or a particular type of data (e.g. MLFriend is designed for event-driven time-series data), as well as by the absence of an user-centric prediction task recommender system, which all lack.

How can we build an end-to-end, automated machine learning pipeline — in our conception, an intelligent data science agent with a autonomy level of 6? In the following sections, we provide a roadmap for future AutoML research that could lead to the development of such a tool, along with important challenges and opportunities that will likely come with achieving this ambitious goal.


\section{Challenges and Opportunities}\label{sec:design}
This section presents the unique challenges of developing a level 6 AutoML agent in terms of design and learning goals. We envision that fully solving these challenges will require more research, and so present some motivating examples in order to inspire the community to jointly pursue them.

\begin{table*}[!htb]\footnotesize
\vspace{-2mm}
\caption{Flight Delay Dataset details}
\vspace{-4mm}
\label{tab:columns}
    {\begin{tabularx}{\linewidth}{@{\extracolsep{\fill}}lX}\toprule
    Attribute Type & Attribute Name and Meanings \\\midrule

    Timestamp       & Date, Day\_of\_week   \\
                    & scheduled\_departure\_hour \\
                     & scheduled\_time, elapsed\_time (the scheduled trip time and the actual time)                     \\\midrule   
                    
    Entities        & Airline, Flight\_number, Tail\_number                                                                                                    \\
                     & Origin\_airport, Destination\_airport                                                         \\\midrule

    Categorical Attributes  &   cancelled\_status,                                         Cancellation\_reason                                      (whether the flight is                                    canceled and reason for cancellation)\\\midrule
                     
    Numerical Attributes    & Departure\_delay, Arrival\_delay (the departure delay and arrival delay in minute)\\
                     & Air\_system\_delay, Security\_delay, Airline\_delay, Late\_aircraft\_delay, Weather\_delay (delay caused by different reasons in minute) \\\bottomrule
    \end{tabularx}}
    \vspace{-5mm}
\end{table*}

\begin{wraptable}{r}{0.5\textwidth}\footnotesize
\vspace{-15pt}
    \caption{Operators}\vspace{-10pt}
    \label{table:op}
    {\begin{tabular}{p{0.5in}p{1.10in}p{0.75in}}
            \toprule
            Operation Set & Supported Ops & Supported Types \\\midrule
            Filter & \texttt{all\_fil} & None\\
            $\mathbf{O}_f$    & \texttt{greater\_fil}, \texttt{less\_fil} & Numerical\\
            & \texttt{eq\_fil}, \texttt{neq\_fil} & Categorical/Entity\\\midrule
            Aggregation & \texttt{count\_agg} & None\\
            $\mathbf{O}_g$    & \texttt{sum\_agg}, \texttt{avg\_agg}, & Numerical \\
               & \texttt{min\_agg}, \texttt{max\_agg} & Numerical \\
            & \texttt{majority\_agg} & Categorical/Entity \\
            \bottomrule
        \end{tabular}}
\end{wraptable}

\subsubsection*{\bf Let's Do a Case Study:}\label{sec:case}
As most real-world data comes in the form of event-driven time series, and the primary goal of a Level 6 AutoML agent is to enable domain experts to perform predictive analysis directly on such data, we focus on event-driven time series data in this case study. We picked a dataset we call \textbf{Flight Delay},\footnote{https://www.kaggle.com/usdot/flight-delays}, which contains flight records for the United States from 2015. As this dataset has a timestamp associated with each event, we pick the time attribute, entity attributes, numerical attributes, and categorical attributes, which we use to generate prediction tasks. For simplicity, we remove unsupported attributes like sets or natural language. Table~\ref{tab:columns} shows the attributes we used to generate possible prediction tasks. Using this data set, we conducted an exploratory case study to understand how a Level 6 AutoML agent could be used in a real-life scenario.

\subsection{\bf Challenges in Task Formulation} \label{sec:PTG}

Our key observation is that across domains, a prediction task is usually characterized by three components: an \emph{outcome function}, the \emph{prediction window} over which this outcome is calculated (denoted \texttt{window}), and the \emph{prediction horizon} over which one wishes to predict this outcome (denoted \texttt{lead}). Abstracting the \pw, \lead and other parameters away from the outcome function allows us to focus on the most important point of translation from human intuition to problem specification, \ie the definition of the outcome function itself. Human data scientists often translate an abstract goal like ``understanding customer engagement'' into a computable quantity that can be predicted. For example, engagement with a particular website could be quantified as \textit{the number of times a customer visits the website during a week}. Here, the outcome is \textit{the number of visits to the website} and the \pw is one week. Prediction task formulation is one of the most critical steps of machine learning; even industry leaders at Facebook have specifically emphasized that: ``\emph{the right set-up is often more important than the choice of algorithm}''~\cite{facebookfieldguide}. What this means for a Level 6 AutoML agent is that we need a formal language to represent prediction tasks accurately, and that this language must be general enough to translate an abstract goal into a computable quantity.

\begin{enumerate}[leftmargin=*]
    \item {\emph {\bf Prediction Task Expression Language:}} To develop a Level 6 AutoML agent, it is necessary to have a language that can be used to specify prediction tasks. We will call this ``Prediction Task Expression Language'', or \textit{PeTEL}. We argue that a complete \textit{PeTEL} needs to successfully express two basic components: 1) a generic \emph{Outcome Function}, and 2) \emph{Search Parameters} (\pw, \texttt{lead} and others) to materialize that outcome function. Previous work, including Trane~\cite{schreck2016would} and MLFriend~\cite{xu2019mlfriend}, has made progress on a reasonable design for Search Parameters, as well as simple expressions for outcome function. However, expressing an outcome function still remains a significant challenge; specifically, how can we express an outcome function that is easily understandable by domain experts with little or no experience in machine learning, so that they can play with it? Fulfilling this goal will require a much larger effort.

As with any language, PeTEL should provide constructs for variables, operators and functions. What should these look like? Variable types can be as simple as integers, real-valued numbers, and categorical attributes, or they can be more complicated variables, like entities and events. The details of these types will vary based on implementation, while the operators will depend on specific variable types. Some basic ones might include filter operators (equal to, greater than, etc.), aggregation operators (count, sum, average, etc.), and smoothing operators (to handle missing values in the data). Table~\ref{table:op} shows some examples of operators for a Level 6 AutoML agent, along with their supported types. For instance, take the following prediction goal:
    
    \begin{center}
    \emph{\color{blue} For each airline, predict the maximum delay suffered by any of its flights.} 
    \end{center}

    This prediction goal can be expressed by the following PeTEL expression: 
    
    \begin{center}
    \begin{minipage}{0.5\textwidth}
        \begin{petel}
        Entity: AIRLINE
        Filter: NONE
        Aggregator: max_agg(<ARRIVAL_DELAY>)
        \end{petel}
    \end{minipage}
    \end{center}
    

    \item{\emph{\bf Interpretability of a Task}:} 
    While PeTEL is about encoding an outcome function into a standard representation, the \emph{interpretability} of a task is essentially the other side of the coin. That is, how can real humans decode/comprehend the actual target outcome from that standard language expression? For example, consider the following PeTEL expression:

   While PeTEL is about encoding an outcome function into a standard representation, the \emph{interpretability} of a task is essentially the other side of the coin: how can humans translate the standard language expression into an actual target outcome? For instance, consider the following PeTEL expression:

    \begin{center}
    \begin{minipage}{0.7\textwidth}
    \begin{petel}
    Entity: AIRLINE
    Filter: eq_filter(<CANCELLATION_REASON, 'bad_weather'>)
    Aggregator: majority_agg(<DESTINATION_AIRPORT>)
    \end{petel}
    \end{minipage}
    \end{center}

    Although people with basic knowledge of structured query language may understand the meaning behind this expression, others will likely have more trouble. However, if a Level 6 AutoML agent translates the PeTEL expression into the following natural language form, it becomes much clearer to the user:

    \begin{center}
    \begin{minipage}{0.7\textwidth}
    \emph{\color{blue}{For each airline, predict which destination airport will have the largest number of flights canceled tomorrow due to inclement weather.}}
    \end{minipage}
    \end{center}

    Thus, translating PeTEL expressions into human-readable language is a core
    challenge that must be addressed in order to create a Level 6 AutoML agent.

\end{enumerate}



\subsection{\bf {Challenges in Prediction Engineering for Identifying ``Promising'' Tasks}}\label{subsec:VIDS:promising-task-identification}

Because the space of possible prediction tasks grows exponentially with the number of attributes (in the dataset) as well as the number of operators (supported by PeTEL); executing full AutoML pipelines for all these tasks is not a computationally feasible option for an interactive Level 6 AutoML agent. One way to address this issue is to filter out uninteresting and invalid tasks, and create a priority list of \textit{promising} tasks that can be evaluated with a feasible amount of computational effort. The primary challenge of identifying \textit{promising} tasks lies in how we define and quantify the \emph{promise} of a prediction task. This is a philosophical question with no single right answer. To simplify things, we identify four metrics which we believe are indicative to some degree of the promise of a prediction task, and discuss the challenges and design choices involved with their computation.


\begin{enumerate}[leftmargin=*]

    \item{\emph{\bf Task Validity}:} While \emph{PeTEL} can generate many different prediction tasks by combining operators and variables, not all combinations will yield a valid task; indeed, many will be invalid. For example, one prediction task for the flight delay dataset can be expressed by the following PeTEL expression: 
    
    \begin{center}
    \begin{minipage}{0.5\textwidth}
    \begin{petel}
    Entity: <AIRLINE>
    Filter: less_filter(<ARRIVAL_TIME>, <DEPARTURE_TIME>)
    Aggregator: count_agg(None)
    \end{petel}
    \end{minipage}
    \end{center}
    
    where, a filter is applied first to check that \mintinline{text}{ARRIVAL_TIME < DEPARTURE_TIME}, and then a count operation is performed over the filtered records after grouping them by airline. Although this is a syntactically correct task, it is not semantically meaningful. Indeed, the natural language translation of the PeTEL expression would be the following:
    
    \begin{center}
    \begin{minipage}{0.5\textwidth}
    \emph{\color{blue} {For each airline, predict how many flights tomorrow will arrive at the destination before they depart.}}
    \end{minipage}
    \end{center}

    This is clearly an invalid task, as it does not make any sense. Thus, one very important challenge for a Level 6 AutoML agent is to figure out how to filter out invalid tasks.

    \item{\emph{\bf User Preference}:} Understanding which sorts of prediction task tend to be preferred by a particular user is crucial for effectively identifying \textit{promising} prediction tasks. Capturing \emph{User Preference} has two major benefits: 1) we can recommend highly relevant task to users and 2) we can easily filter out tasks the user dislikes, saving time and computational resources. However, modeling \emph{User Preference} in this context is tricky, and often specific to a particular user and their domain. For example, an analyst at an airport authority may be more interested in predicting whether their airport will have an unusual number of delayed flights so that they can accommodate passengers, while an analyst at an airline may be more interested in whether their own flights will be delayed. Although both tasks are valid, they may not be equally interesting to different users. Thus, the effectiveness of filtering \textit{promising} tasks for a user depends heavily on accurate modeling of that user's preferences. In the absence of a pre-trained user model, an active learning method for capturing user preferences, such as the \emph{Machine Teaching} perspective introduced by Microsoft, can become really handy~\cite{simard2017machine}.

    \item{\emph{\bf Potential Business Value}:} Another metric for filtering out less \textit{promising} tasks involves quantifying their potential business value and discarding tasks with low scores. Assuming that a particular prediction task can achieve reasonably high accuracy, would it  impact the business significantly? If not, it may not be worth the effort to train a machine learning model to carry it out. \emph{Potential Business Value} is often very hard to quantify, and the best option may be to ask domain experts with long-term experience to foresee the impacts of such models.

    \item{\emph{\bf Sufficient Training Examples}:} Given a valid and highly interesting prediction task, another important metric related to \textit{promise} is the number of training examples it’s possible to extract from the available data. If this number is very low, it may not be very meaningful to train a model on such a small corpus. In the case of classification tasks, one must also ensure that enough training examples from each class are present in the training data.

\end{enumerate}

One can compute the overall promise of a prediction task by combining these four metrics: User Preference, Task Validity, Business Potential Value, and Sufficient Training Examples. Again, this is not an absolute standard by any means, and this remains an open problem.

\vspace{-2mm}

\subsection{Challenges in Result Summary and Recommendation (RSR)}\label{subsec:VIDS:design-of-prediction-task-recommendation}\vspace{-1mm}

As mentioned in section~\ref{sec:pipeline}, \textit{Result Summary and Recommendation} (RSR) is a data scientist’s most important job, and is still done largely manually, without any systematic structure. To realize a Level 6 AutoML agent, even this must be done automatically. The goal here is to rank prediction tasks based on their utility scores, and to present the top-k prediction tasks to domain experts or other users. Below we present a brief summary of major challenges associated with the automation of the \emph{RSR} sub-task.

\vspace{-1mm}
\subsubsection{\bf Evaluation Challenges: }  Prediction task recommendation essentially boils down to ranking tasks based on their utility to the user. But how should we evaluate the utility of a prediction task, and what does ``utility'' really mean in this context? In other words, how do we evaluate the performance of a prediction task recommender system? We hypothesize that utility scores can be computed as a function of the task-specific metrics discussed in section~\ref{subsec:VIDS:promising-task-identification} including pre-AutoML metrics like \emph{User Preference}, \emph{Task Validity}, etc. as well as post-AutoML metrics like \emph{Prediction Accuracy}, \emph{Computation Time}, etc. Some of these metrics can be computed automatically, like \emph{Sufficient Training Examples} and \emph{Prediction Accuracy}, while some metrics, like \emph{User Preference} and \emph{Potential Business Value}, are highly dependent on actual users — in this case domain experts or business leaders. Thus, utility appears to be a complex metric composed of simpler metrics, some of which are computed automatically and some of which are collected through human annotations and feedback. However, it is unclear which types of user feedback — user rating, pairwise comparison, partial ranking, etc. --- are the most appropriate for gathering annotations.

\vspace{-1mm}
\subsubsection{\bf Representation Challenges: } The next challenge in designing a prediction task recommender system is finding a way to represent prediction tasks for the machine learning setup. This involves transforming the prediction task descriptions into a feature vector. While the pre-AutoML and post-AutoML metrics can certainly be used to compute meaningful numerical features, what about features related to task description? How do we convert the PeTEL expressions into a set of features suitable for machine learning? Should we choose a binary vector over the space of the attribute set, aggregation and filter operations? Should we also consider the natural language descriptions through some simple language models, like \emph{unigram} or \emph{bigram}? Or, given the promise of neural architectures, should we build a distributed representation or embedding for prediction task featurization? A detailed investigation of these questions will be required to reach a conclusion.

\begin{table*}[!htb]\footnotesize

    \caption{Example prediction tasks on the \emph{Flight Delay}  data set. {\bf PeTEL} is the structured representation of the problem, including the entity, attributes and operations. {\bf Intermediate} is an automatically generated natural language description of the problem. {\bf Natural} is the human interpretation of the problem.} \label{table:example}
    \vspace{-2mm}
    \begin{tabular*}{\linewidth}{|p{\linewidth - \tabcolsep * 2}|}
    
    \hline
    
    {\bf PeTEL}:
    \mintinline[formatcom=\color{mymaroon},fontsize=\small]{text}{Entity: AIRLINE, Filter: NONE, Aggregator: max_agg(<ARRIVAL_DELAY>)} \\
    {\bf Intermediate}: \color{myOliveGreen}{For each $<$AIRLINE$>$ predict the maximum $<$ARRIVAL\_DELAY$>$ in all related records}\\
    {\bf Natural}: \color{blue}{Predict the delay time of each airline's most delayed flight tomorrow}\\
    
    \hline
    
    {\bf PeTEL}:  \mintinline[formatcom=\color{mymaroon},fontsize=\small]{text}{Entity: AIRLINE, Filter: less_filter(<ELAPSED_TIME>), Aggregator: }\\ \mintinline[formatcom=\color{mymaroon},fontsize=\small]{text}{sum_agg(<AIRLINE_DELAY>)}\\
    {\bf Intermediate}: \color{myOliveGreen}{For each $<$AIRLINE$>$ predict the total $<$AIRLINE\_DELAY$>$ in all related records with $<$ELAPSED\_TIME$>$ less than \_\_}\\
    {\bf Natural}: \color{blue}{Predict the total flight delay tomorrow for each airline's short-haul flights}\\
    
    \hline
    
    {\bf PeTEL}: \mintinline[formatcom=\color{mymaroon},fontsize=\small]{text}{Entity: AIRLINE,  Filter: greater_filter(<AIRLINE_DELAY>), Aggregator: } \\ \mintinline[formatcom=\color{mymaroon},fontsize=\small]{text}{majority_agg(<DESTINATION_AIRPORT>)}\\
    {\bf Intermediate}: \color{myOliveGreen}{For each $<$AIRLINE$>$ predict the majority of $<$DESTINATION\_AIRPORT$>$ in all related records with $<$AIRLINE\_DELAY$>$ greater than \_\_}\\
    {\bf Natural}: \color{blue}{For each airline, predict which destination is the most likely to have delays caused by the airline itself} \\
    
    \hline
    
    {\bf PeTEL}: \mintinline[formatcom=\color{mymaroon},fontsize=\small]{text}{Entity: ORIGIN_AIRPORT, Filter:  greater_filter(<DEPARTURE_DELAY>), Aggregator: } \\ \mintinline[formatcom=\color{mymaroon},fontsize=\small]{text}{majority_agg(NONE)}\\
    {\bf Intermediate}: \color{myOliveGreen}{For each $<$ORIGIN\_AIRPORT$>$ predict the number of records with $<$DEPARTURE\_DELAY$>$ greater than \_\_}\\
    {\bf Natural}: \color{blue}{Predict how many flights departing from this airport will be delayed more than 1 hour}\\
    
    \hline
    
    {\bf PeTEL}: \mintinline[formatcom=\color{mymaroon},fontsize=\small]{text}{Entity: AIRLINE, Filter:  eq_filter(<CANCELLATION_REASON>), Aggregator: } \\ \mintinline[formatcom=\color{mymaroon},fontsize=\small]{text}{majority_agg(<DESTINATION_AIRPORT>)}\\
    {\bf Intermediate}: \color{myOliveGreen}{For each $<$AIRLINE$>$ predict the majority of $<$DESTINATION\_AIRPORT$>$ in all related records with $<$CANCELLATION\_REASON$>$ equal to \_\_}\\
    {\bf Natural}: \color{blue}{Predict which destination airport has the largest number of flights canceled tomorrow due to weather} \\
    
    \hline
    
    \end{tabular*}
    \vspace{-5mm}
\end{table*}

\vspace{-1mm}
\subsubsection{\bf Learning challenges: } Given a particular \emph{utility} metric and feature representation for prediction tasks, a crucial challenge is finding the best way to learn this \emph{utility} metric. As with any learning system, this comes with basic challenges including the following: How do we create an effective training corpus? How should we model the ranking function of the recommender system? What is the ideal objective function in this case?

As a starting point, we identify four different ways to rank prediction tasks in order to provide users with recommendations. Below, we provide a brief summary of each.

\begin{enumerate}[leftmargin=*]
  \item{\emph{\bf Static Ranking:}} The simplest way to recommend prediction tasks is to rank them according to their \emph{utility} score, and present tasks with the highest scores. Through initial experiments conducted on the \emph{Flight Delay} dataset, we present some examples of highly recommended tasks in Table~\ref{table:example}, which we curated through manual inspection. This static ranking list can be presented with or without each tasks’ \emph{utility} score; however, it is currently unclear whether user experience will vary based on whether this \emph{utility} score is shown or not.
  
  Static Ranking is appropriate when ranking is based on well-defined numerical scores that can be computed automatically. For example, if we want the ranker to show tasks with high prediction accuracy only, then static ranking is sufficient for our purposes. Static Ranking may also work in cases where we have a very robust user model that has already been learned on a large amount of historical data. For example, if we know the preferences of a particular user very well, Static Ranking may work for that user.

  \item{\emph{\bf Interactive Ranking with Feedback}:} This setup is required in cases when the \emph{utility} function depends heavily on the user and a robust model for user preference is not available beforehand. In this setup, a Level 6 AutoML agent can first recommend a set of tasks that have been scored highly by the recommender system, and ask the user to label these tasks as ``useful'' and ``not-useful.'' By looking at this feedback, a Level 6 AutoML agent can actively learn the user's preferences and revise the recommendation model to accommodate them (see ~\cite{long2014active} for details on how active learning can be used for ranking tasks). This recommendation-modification loop can continue until the user is satisfied.

  \item{\emph{\bf Recommendations from previous experiences}:}
  This setup is useful as a starting point in cases when the problem domain is not known historically. For example, a recommender system may have been previously trained on car rental data, but not on truck rental data; these two rental systems may share similarities and differences the specifics of which are unknown. One reasonable way to approach this problem is to transfer knowledge from the car rental domain and use it for initial recommendations in the truck rental domain, an increasingly popular strategy known as transfer learning~\cite{marcelino2018transfer}.

  \item{\emph{\bf Combining Interactive Ranking with Transfer Learning}:} A more sophisticated way of recommending would be to start with a ranker that’s been learned on a similar domain with sufficient training examples, and interactively update that ranker by getting feedback from users of the target domain. Going back to the car vs. truck rental example, one can start with the recommender model learned from the car rental domain to provide recommendations in the truck rental domain while gathering feedback from users about the quality of the recommendations. Once the system gathers more information about the truck rental domain, the recommender model can be revised to capture this new information and better serve the target domain.
\end{enumerate}

\subsubsection{\bf Diversity challenges:} Another challenge for the recommender system is to ensure that the list of recommendations includes a diverse set of tasks. For example, consider the following prediction tasks:

\vspace{-3mm}
\begin{table}[h!]
        \centering
         \begin{tabular}{l l} 
         {\bf Task 1:} predict the maximum $weather\_delay$ & {\bf Task 3:} predict the average $security\_delay$\\
         {\bf Task 2:} predict the average $weather\_delay$ & {\bf Task 4:} predict the total $security\_delay$\\
         \end{tabular}
         \vspace{-3mm}
\end{table}

If we were to recommend two prediction tasks from this set of four, which two would ensure more diversity in terms of attributes and operations? Task 1 and 2 predict $weather\_delay$, while, Task 3 and 4 predict $security\_delay$. Clearly, task-set $\{1,3\}$ is more diverse than task-set $\{1,2\}$, while, task-set $\{2,3\}$ is less diverse than task-set $\{2,4\}$. However, the degree of diversity required depends on the user's preference, and there is not yet any clear, general standard for evaluating it. 


\section{Human-AI interaction in level 6 AutoML agent} \label{sec:Human}

Ultimately, human analysts and decision-makers must be convinced that prediction tasks recommended by the Level 6 AutoML agent are indeed useful and interesting. This goal brings up several interesting challenges and opportunities in Human-AI interaction, some of which we highlight below. 

\begin{itemize}[leftmargin=*]

  \item {\bf Problem Space Understanding:} Human analysts need to have some understanding of the problem space — the set of all prediction tasks that could possibly be generated by the Level 6 AutoML agent — in order to grasp the context in which the data science process takes place. This closely relates to the problems of looking at raw data and identifying useful tables and attributes, understanding the frequency at which the data is collected and made available, and so on. The domain expert should recognize that the Level 6 AutoML agent can enumerate only a subset of this infinitely large problem space.

  \item {\bf Prediction Task Understanding:} When the Level 6 AutoML agent presents a prediction task to a human analyst, what form should this presentation take? Multiple interfaces are possible: a raw PeTEL expression, a basic generated natural language description, a more sophisticated natural language explanation, a visual representation of the task in terms of inputs, a visualization of prediction instances associated with that task, or something else entirely. Such decisions are quite important, as true understanding and accurate evaluation of a given prediction task may depend on how it is presented to the user.

  \item {\bf Prediction Task Evaluation:} As described in Sections~\ref{subsec:VIDS:promising-task-identification} and \ref{subsec:VIDS:design-of-prediction-task-recommendation}, there are many dimensions along which a prediction task can be evaluated. Human evaluators must be able to judge performance along each of these dimensions, often making a difficult connection between the prediction task at hand and other processes or outputs within their organization.

  \item {\bf Prediction Task Operationalization:} After the selection of prediction tasks, the job is in some ways just beginning. End-to-end machine learning pipelines must be developed and deployed to solve these tasks. This involves additional challenges, such as setting any free parameters for the task, or adjusting prediction windows. These may involve additional rounds of communication between the human analyst and the Level 6 AutoML agent — raising similar questions to those delineated above, as well as new ones. 

\end{itemize}

\vspace{-2mm}
\section{The Long Road Ahead}\label{sec:conclusion}
The AutoML community has provided us with powerful tools for executing machine learning tasks and producing predictive models. 
However, defining prediction problems themselves and navigating the enormous space of alternatives still poses significant challenges even for data scientists — let alone for domain experts and business leaders, who still struggle to directly access the power of big data analytics. Indeed, a significant portion of the machine learning process today requires involvement from a data scientist, making the whole process inefficient and inaccessible to a wider audience. 
In this review paper, we explained the gravity of this problem by introducing a new taxonomy for AutoML solutions consisting of seven distinct autonomy levels, and designed to highlight gaps and limitations in current machine learning practice. We then presented our vision for a more intelligent agent (Level 6) to address this issue. 

A Level 6 agent would provide two main benefits: 1) It would increase the productivity of data scientists by automating prediction task formulations and recommendations, and 2) It would enable domain experts to directly access and interact with the fascinating world of predictive machine learning, opening up new opportunities for them to apply this powerful tool to real-world problems. With this in mind, we make the following key observations. First, formulating a prediction problem is challenging, and there is currently no established standard for accomplishing this task systematically. Second, the key technical challenge of realizing a Level 6 AutoML agent boils down to designing an interactive prediction task recommender system capable of actively engaging the user.

We presented an exploratory case study using a real-life dataset to demonstrate the feasibility of actually realizing the currently hypothetical Level 6 AutoML agent. Our initial exploration reveals the many design and technical challenges that must be addressed in order to reach the eventual goal of automated data science — in particular automated task formulation, effective prediction engineering and the recommendation of useful tasks. Another crucial but orthogonal issue is the evaluation of machine learning models, as fair evaluation is critical to the success of AutoML. Indeed, the fair evaluation of ML models depends on accurate design choices made throughout the entire \textit{data science} process:  training and testing datasets must be created in such a way as to ensure an accurate and robust model is learned; benchmarking datasets and evaluation test beds must be built in order to precisely and reliably assess the model’s accuracy and performance; training labels should be created in a way that ensures minimal annotation bias in the ML models; etc. In summary, it is very important that a Level 6 AutoML tool ensures an unbiased and representative model evaluation, as this is critical to the success of AutoML in general. 


We view a Level 6 AutoML agent as a tool that will increase the productivity of data scientists and domain experts. With this paper, we aim to draw the attention of the broader data science and human-computer interaction community, and urge them to invest significant research efforts in this direction. The long road ahead demands an exciting interdisciplinary research approach involving multiple areas of computer science, especially human-computer interactions, information retrieval, databases,  programming language design, and software engineering.


\bibliographystyle{ACM-Reference-Format-Journals}
\bibliography{bibs4cacm,Lei}

\section*{Online Resources}

\begin{urlist}
\item\label{OpenRefine}\url{http://openrefine.org/}
\item\label{TrifactaWrangler}\url{https://www.trifacta.com/products/wrangler/}
\item\label{Drake}\url{https://github.com/Factual/drake}
\item\label{TIBCOClarity}\url{https://clarity.cloud.tibco.com/landing/feature-summary.html}
\item\label{DataQuality}\url{https://sourceforge.net/projects/dataquality/})
\item\label{Winpure}\url{https://winpure.com/}
\item\label{TalentedDataQuality}\url{https://www.trustradius.com/products/talend-data-quality/reviews}
\item\label{DataLadder}\url{https://dataladder.com/}
\item\label{DataCleaner}\url{https://datacleaner.org/}
\item\label{Cloudingo}\url{https://cloudingo.com/}
\item\label{Refier}\url{http://nubetech.co/technology/}
\item\label{IBMInfosphereQualityStage}\url{https://www.ibm.com/uk-en/marketplace/infosphere-qualitystage}
\item\label{FeatureTools}\url{https://www.featuretools.com/}
\item\label{featexp}\url{https://github.com/abhayspawar/featexp}
\item\label{MLFeatureSelection}\url{https://github.com/duxuhao/Feature-Selection}
\item\label{FeatureEngineeringFeatureSelection}\url{https://github.com/Yimeng-Zhang/feature-engineering-and-feature-selection}
\item\label{FeatureHub}\url{https://github.com/HDI-Project/FeatureHub}
\item\label{featuretoolsR}\url{https://github.com/magnusfurugard/featuretoolsR}
\item\label{Feast}\url{https://cloud.google.com/blog/products/ai-machine-learning/introducing-feast-an-open-source-feature-store-for-machine-learning}
\item\label{ExploreKit}\url{https://github.com/giladkatz/ExploreKit}
\item\label{HyperparameterHunter}\url{https://github.com/HunterMcGushion/hyperparameter_hunter}
\item\label{AmazonSagemaker}\url{https://aws.amazon.com/sagemaker/}
\item\label{hyperband}\url{https://github.com/zygmuntz/hyperband}
\item\label{BigMLOptiML}\url{https://bigml.com/api/optimls}
\item\label{Hyperboard}\url{https://github.com/WarBean/hyperboard}
\item\label{GoogleHyperTune}\url{https://cloud.google.com/ml-engine/docs/using-hyperparameter-tuning}
\item\label{SHERPA}\url{https://github.com/sherpa-ai/sherpa}
\item\label{IndieSolver}\url{https://indiesolver.com/}
\item\label{Milano}\url{https://github.com/NVIDIA/Milano}
\item\label{MindFoundryOPTaaS}\url{https://www.mindfoundry.ai/mind-foundry-optimize}
\item\label{BBopt}\url{https://github.com/evhub/bbopt}
\item\label{sigopt}\url{https://sigopt.com/}
\item\label{adatune}\url{https://github.com/awslabs/adatune}
\item\label{gentun}\url{https://github.com/gmontamat/gentun}
\item\label{optuna}\url{https://github.com/optuna/optuna}
\item\label{test-tube}\url{https://github.com/williamFalcon/test-tube}
\item\label{Advisor}\url{https://github.com/tobegit3hub/advisor}
\item\label{DotData}\url{https://dotdata.com/}
\item\label{TPOT}\url{https://github.com/EpistasisLab/tpot}
\item\label{IBMAutoAI}\url{www.ibm.com/Watson-Studio/AutoAI}
\item\label{Auto-Sklearn}\url{https://github.com/automl/auto-sklearn}
\item\label{AzureMachineLearning}\url{https://azure.microsoft.com/en-us/services/machine-learning/}
\item\label{GoogleCouldAIplatform}\url{https://cloud.google.com/ai-platform/}
\item\label{AmazoonAWSservice}\url{https://aws.amazon.com/marketplace/solutions/machine-learning/data-science-tools}
\item\label{DataRobot}\url{https://www.datarobot.com/}
\item\label{AmazonForecast}\url{https://aws.amazon.com/forecast/}
\item\label{AutoKeras}\url{https://github.com/keras-team/autokeras}
\item\label{H2ODriverlessAI}\url{http://docs.h2o.ai/driverless-ai/latest-stable/docs/userguide/index.html}
\item\label{AdaNet}\url{https://github.com/tensorflow/adanet}
\item\label{CloudAutoML}\url{https://cloud.google.com/automl/}
\item\label{PocketFlow}\url{https://github.com/Tencent/PocketFlow}
\item\label{C3AISuite}\url{www.c3.ai}
\item\label{automl-gs}\url{https://github.com/minimaxir/automl-gs}
\item\label{FireFly}\url{www.firefly.ai}
\item\label{MLBox}\url{https://github.com/AxeldeRomblay/MLBox}
\item\label{Builton}\url{www.builton.dev/ml-apis}
\item\label{Morph-net}\url{https://github.com/google-research/morph-net}
\item\label{DeterminedAIPlatform}\url{www.determined.ai}
\item\label{ATM}\url{https://github.com/HDI-Project/ATM}
\item\label{TransmogrifAI}\url{https://github.com/salesforce/TransmogrifAI}
\item\label{RemixAutoML}\url{https://github.com/AdrianAntico/RemixAutoML}
\item\label{OpenML100}\url{https://www.openml.org/s/14}
\item\label{OpenML-CC18}\url{https://www.openml.org/s/98}
\item\label{AutoMLChallenges}\url{http://automl.chalearn.org/data}
\end{urlist}

\newpage

\appendix

\section{Automation in Data visualization, Cleaning and Curation (DCC)}\label{appen:DCC}

\subsection{\textbf{Research Contributions}}

{\bf Framework Based Approach:} One school of researchers primarily adopted a framework-based approach for data visualization, cleaning, and curation. For example, \citet{miao2018assessment} introduced a data quality assessment and cleaning framework specifically for electronic health records (EHR). They used this framework to conduct a case study to estimate the risk factors for hip fracture patients who might need to be readmitted within one month of their surgery. During this case study, they identified 23 critical data quality problems, 16 of which were addressed by the proposed framework. In a similar effort, \citet{zhang2017time} proposed a new framework called \textit{Iterative Minimum Repairing} (IMR) to iteratively repair anomalies in time series data. In each iteration, IMR tries to identify the most confident anomaly and repair it; thus, it minimally changes one point at a time. The authors have also designed an incremental algorithm for parameter estimation, which can reduce the complexity of parameter estimation from $O(n)$ to $O(1)$. In order to enhance the accuracy of repairs in later iterations, the authors tried to learn the temporal nature of errors in anomaly detection. 
The paper also demonstrated that IMR method's performance is significantly better than other state-of-the-art approaches, including simple anomaly detection and constraint-based repairing.
 

{\bf Category Based Approach:} \citet{minnich2016clearview} introduced a cleaning process (called ``ClearView'') for online review mining purposes by categorizing noise into three categories. First is the syntactic noise, usually misspelled words, on which they perform two types of cleaning: character-level and word-level. The second is semantic noise, for which they analyze the structure of a sentence. More specifically, they used Stanford CoreNLP Parser and utilized the confidence score from the parser as a semantic correctness measure of a sentence. The third is rating noise, which refers to the discrepancy between valid text sentiment and star rating. They used an ensemble of sentiment classifiers to iteratively label the sentiment of a review to maximize the agreement with the star-rating provided by the user. 

\citet{cheng2018data} took an indicator based approach for data quality assessment and associated cleaning strategy. They introduced four indicators for data quality assessment in wireless sensor networks - amount of data, correctness, completeness, and time correlation index measure - and studied the outcomes of different orders of cleaning strategy. They also proposed a strategy to avoid redundant cleaning operations and reduce cleaning expenses while ensuring high data quality. This strategy starts with computing the volume indicator of the data-set and determine if the cleaning process is necessary at all. If it is necessary, they then execute the cleaning process via completeness, time-related, and correctness indicators.

{\bf Active Approach:} \citet{krishnan2016activeclean} used an active cleaning strategy while preserving convergence guarantees in the context of statistical model training. They proposed a new technique called `ActiveClean' which supports convex loss models while prioritizing data with result alteration potential. Specifically, \textit{ActiveClean} suggests a sample of data to clean based on the data's value to the model and the likelihood that it is dirty. \textit{ActiveClean} interleaves cleaning and training process iteratively by starting with a dirty model as initialization and taking gradient steps (cleaning a sample  of records) towards a global solution (a clean model). For the same amount of data cleaned, proposed optimizations in ActiveClean can improve model accuracy by up to 2.5x. In comparison with techniques such as \textit{Active Learning} and \textit{SampleClean} which are not optimized for sparse low-budget setting, \textit{ActiveClean} achieves model with high accuracy where it cleans far fewer records.

{\bf End-to-end Systems:} End-to-end cleaning systems have been recently studied by some researchers. \citet{chu2015katara} presented an end-to-end data cleaning system named `KATARA' that uses trustworthy knowledge bases (KB) and crowdsourcing for data cleaning. Given a table, KB and crowd, KATARA interprets the table semantics with respect to KB, identifies wrong data, and recommends top-k possible repairs. A graphical interface is provided to assist users in tuning the parameters. Given a selected table and KB, KATARA will cross-validate each -tuple in the table with KB. If a -tuple is inconsistent with KB, an ambiguity is raised. The ambiguity is then presented to the crowd to get feedback, which can help KATARA repair the data more accurately. 

Another end-to-end data cleaning workflow is ``Wisteria'' proposed by \citet{haas2015wisteria}. \textit{Wisteria} allows users to specify data cleaning plans and allows iterative revision of the plan. On each iteration, \textit{Wisteria} presents a cost-aware recommendation to improve the accuracy of the plan by swapping in a new cleaning operator.

\subsection{\textbf{Open Source and Commercial Systems}}
There are several open-source and commercial systems publicly available for performing data visualization, cleaning, and curation. Below we present brief highlights of these systems.

\begin{itemize}[leftmargin=*]
    \item \textbf{Open Refine} \uref{OpenRefine} allows users to clean and transform their data through a browser-based interface. This is installed in a client's machine; thus clients do not need to share their data unless they are contributing to the software. This system cleans data based on users decision. 

    \item \textbf{Trifacta Wrangler}\uref{TrifactaWrangler} is a free cloud service that organizes and structures data automatically as soon as it is imported into the system. The system then suggests transformation and aggregation. Within a managed cloud platform, a team can share their work, schedule a workflow, and connect more data. 

    \item \textbf{Drake} \uref{Drake} is a text-based work-flow tool which automatically resolves data dependency and calculates command list and order. It has HDFS support for multiple input and outputs and it includes host features to bring sanity to chaotic data processing workflows. 

    \item \textbf{TIBCO Clarity} \uref{TIBCOClarity} is an on-demand data preparation tool. It detects data types and pattern which is later used for auto-metadata generation. It categorizes data using some predefined facets on text patterns. It also provides numeric distribution for variance identification within the data and a configurable fuzzy matching algorithm for duplicate detection.

    \item \textbf{Open Source Data Quality and Profiling} \uref{DataQuality} is an integrated high performing data management platform. It also has Hadoop support to move files from a Hadoop grid and create Hive tables.

    \item \textbf{Winpure} \uref{Winpure} provides a data cleaning product which automatically determines data impurities and suggests solutions about different data quality problems. It offers users different data transformations and data preparation operations through various templates. It also allows user to load data from different types of data sources and save transformed data in various data types.

    \item \textbf{Talented Data Quality}  \uref{TalentedDataQuality} analyzes raw data using match pairing component which uncovers and labels suspicious data. Using those data, they build a classification model which predicts matches for new data sources.
 
    \item \textbf{Data Ladder} \uref{DataLadder} provides a highly visual data cleaning application which automatically profiles data from the start. They use pattern recognition and fuzzy logic for duplicate reduction and cleaning unstructured data. They assure data quality by providing a firewall that prevents bad data from third party through APIs. 
 
    \item \textbf{Data Cleaner} \uref{DataCleaner} includes three different types of operations: analyzers, transformers and filters. A user can create jobs and the system will show him/her a preview of the changes ordered before executing the job. The user can save the current job for future usages. They also provide a web application for monitoring and sharing analysis jobs and results.
  
    \item \textbf{Cloudingo} \uref{Cloudingo} enables users to create a set of filters that will be used to remove duplicate entries from data-sets. Users can pass the data through one or multiple filters that they defined and execute them manually or automatically. It also allows user to schedule a filter job that will curate data at regular intervals. When fetching data, cloudingo automatically runs base filter job to remove duplicates within existing data and new data.
  
    \item \textbf{Refier} \uref{Refier} automatically curates data by learning string similarity and deduce the optimal fuzzy matching rules for any domain or language. Refier needs a training set to learn about the rules, but if unavailable, then need to provide a small set of matching rules. Reifier uses ``Apache Spark'' for both distributed and standalone data.
  
    \item \textbf{IBM Infosphere Quality Stage} \uref{IBMInfosphereQualityStage} enables users to investigate, clean and manage their data. It uses both built in and user defined data quality rules to curate data automatically. It standardizes the data coming from different sources into a common format by removing duplicates and merging multiple systems into a single view. 

\end{itemize}

\section{Automation in Feature Engineering (FE)}\label{appen:FE}

\subsection {\textbf{Research Contributions}}

{\bf{Transformation Based Approach:}}
One popular way of automating feature engineering is to apply different transformations on original features to synthesize new features. For example, \citet{katz2016explorekit}  have identified some basic operators which can transform a feature or combine several to create a new feature. The intuition behind this transformation based approach is the following: ``highly informative feature often results from manipulations of elementary ones.'' Using these operators, \citet{katz2016explorekit} synthesized multiple candidate features which were fed into the training process, producing a candidate feature which is added to the model based on its empirical performance during the training. They have conducted extensive evaluation on 25 data-sets and three classification algorithms and shown that they can achieve overall 20\% classification-error reduction. 

Being inspired by \citet{katz2016explorekit}, \citet{van2017automatic} used three different categories of operators (unary, binary, and group-by-them) to transform old features into new ones. They also used a background classifier to predict the usefulness of a feature. Another contribution of \citet{van2017automatic} is to separate the feature generation and feature selection steps. As a result, feature selection can now be performed on both original and derived features, which was not possible in previous approaches.

{\bf{Generation Based Approach:}}
\citet{kanter2015deep} developed a Deep Feature Synthesis algorithm that can automatically generate features for relational data-sets. Their algorithm follows relationships in the data to a base field, and then sequentially applies mathematical functions along that path to create the final feature. Given entities, their data tables, and relationships, they defined a number of mathematical functions that are applied to create features at two both the entity and relational levels. They optimized the whole pathway by implementing an autotuning process to help it generalize to different data-sets.  

\citet{mountantonakis2017linked} proposed a generative method based on linked data for discovering, creating, and easily selecting valuable features to describe a set of entities in any machine learning (ML) problem. Linked Data refers to a method of publishing structured data while its ultimate objective is linking and integration. In this process, they first discover data-sets and URIs that contain information for a set of entities. Then, they provide users a large number of potential features that can be created for these entities. Finally, they  automatically produce a data-set for the features which were selected by the user. They evaluated this approach by performing a 5-fold cross validation for estimating the performance of different models for the produced data-sets.

{\bf{Learning Based Approach:}} Some researchers have framed feature generation as a learning task and trained ML models to generate informative features. \citet{khurana2017feature} presented such an approach using reinforcement learning (RL) to automate feature engineering (FE)). Their strategy is to train an agent on FE examples, enabling the agent to learn an effective strategy to explore available FE choices under a certain budget constraint. This exploration is performed on a directed acyclic graph called ``transformation graph,''  representing relationships among different transformed versions of the data. Across a variety of data-sets, they have shown that models produced by their system reduced median error rate by 25\%.

\citet{kaul2017autolearn} presented a regression based feature learning algorithm called ``AutoLearn''. They apply regression on each feature to predict the values of other feature and use this regression forecast to supplement additional information in each record. \textit{AutoLearn} includes four major steps. First, prepossessing based on information gain is done to filter out less informative features. Second, using distance correlation, pair-wise correlated features are defined and searched in the original feature space. Third, the original feature space is transformed into the new feature space by learning a predictive relation between correlated features. Finally, newly constructed feature space is constrained by selecting a subset of features based on stability and information gain.  The authors have shown that, compared to original feature space, features learned through their model can improve prediction accuracy by 13.28\% and 5.87\% with respect to other top performing models.

\subsection{\textbf{Open Source and Commercial Systems}}

\begin{itemize}[leftmargin=*]
    \item \textbf{ExploreKit} \uref{ExploreKit} is an automated feature generation kit which is essentially an implementation of the transformation based approach introduced by \citet{katz2016explorekit}.
    
    \item \textbf{FeatureTools} \uref{FeatureTools} is an automated feature engineering framework based on the Deep Feature Synthesis method developed by \citet{kanter2015deep}. \textit{FeatureTools} can effectively perform feature engineering for relational and temporal data. They also provide an R implementation named featuretoolsR \uref{featuretoolsR}.
    
    \item \textbf{Featexp} \uref{featexp} is a python package for feature exploration that shows graphs and plots to help users better understand the feature space. Featexp bins a feature into equal population bins and shows the mean value of the dependent variable (target) in each bin, the trend of which can help users understand the relationship between the target and the feature. Users can compare feature trends to identify noisy features by looking at the number of trend direction changes and the correlation between train and test trend. These graphs / plots also help user to do feature debugging and leakage detection.
    
    \item \textbf{MLFeatureSelection} \uref{MLFeatureSelection} has broken feature selection into three parts. First, it selects a sequence from raw data which will potentially contain the best feature combination. Second, it creates feature groups from the raw data and iteratively update them and remove less important features. Eventually, this process yields a set of important features. Third, a coherent threshold is picked and all the features having higher coherence value are selected as best possible features.
    
    \item \textbf{Feature Engineering \& Feature Selection} \uref{FeatureEngineeringFeatureSelection} provides a guideline for feature engineering and feature selection techniques including why to use them, when to use them and when to use which algorithm.
    
    \item \textbf{FeatureHub} \uref{FeatureHub} provided a collaborative framework for feature engineering and instantiate this idea into a platform. This platform administers collaboration among data scientists where users can register features in the database as well as reuse features written by other users. FeatureHub automatically scores the features so users can get real-time feedback for their features.
    
    \item \textbf{Feast} \uref{Feast} introduces a centralized feature store which can work as a foundation of features used by different projects. Feast allows users to use historical feature data to produce sets of features for training ML models, and makes feature data available to models in production by serving API and ensuring consistency between training and serving is preserved.
\end{itemize}

\section{Automation in Alternative Models Exploration, Testing and Validation (ATV)}\label{appen:ATV}

\subsection{Automation in Hyperparameter Tuning}\label{appen:HT}
\subsubsection {\bf Research Contributions}
There is a vast amount of literature regarding the automation of hyperparameter optimization within a machine learning model. Initially, grid search was tried to explore the space of hyperparameters but researchers immediately realized that the search space gets too large to find a good set of hyperparameters within a reasonable amount of time. Interestingly, it was found that random search often does a decent job for finding an optimal set of hyperparameters~(\citet{bergstra2012random}). Afterwards,  researchers have tried sequential model based optimization (SMBO) by utilizing computer clusters and GPU processors (\citet{bergstra2011algorithms}), and removing SMBO limitations (\citet{hutter2011sequential}). Recent research shows that Bayesian optimization framework used for hyperparameter tuning provides promising results (\citet{snoek2012practical}). Some researchers have also tried reversed stochastic gradient descent with momentum (\citet{maclaurin2015gradient}) which yields promising results too. In this section, we will briefly review the research landscape of automating hyperparameter optimization.

{\bf{Random Search Based Approach:}}
\citet{bergstra2012random} have shown that random search proves to be more efficient for hyperparameter optimization than grid search and manual search. Over the same domain, random search is able to find good or better models with a fraction of the computation time needed for pure grid search. They have also shown that for most data sets, only a handful of hyperparameters matter; but at the same time, the hyperparameters that actually matter are different for different data sets. Grid search suffers from poor coverage of dimensions that really matter, as it allocates too many trials to explore dimensions that matter less. In most cases, random search finds a better model while consuming less computational time.

{\bf{Sequential Model Based Approach:}}
\citet{bergstra2011algorithms} have shown that, with the current availability of powerful computer clusters and GPU processors, it is possible to run more trials. As a result, using an algorithmic approach to optimize hyperparameters can find better results than random search. They have introduced two greedy sequential methods to optimize hyperparameter for neural networks and deep belief networks (DBN), and shown that random search obtains unreliable results for training DBNs. Their two approaches for sequential model-based optimization are: Gaussian Process (GP) approach and Tree-structured Parzen Estimator (TPE) approach. 

\citet{hutter2011sequential} identified three key limitations of sequential model based optimization (SMBO) that prevents it from being used for general algorithm configuration task. The limitations they defined are: (1) SMBO only supports numerical parameter; (2) target algorithm performance is optimized by SMBO only for single instances; and (3) SMBO lacks mechanisms for terminating poorly performing target algorithm early. This work attempts to remove the first two of these SMBO limitations. Through the generalization of SMBO framework, they introduced four components and based on them, they defined two novel SMBO instantiations which are: Random Online Adaptive Racing (ROAR) procedure and Sequential Model-based Algorithm Configuration (SMAC). ROAR is a simple model-free instantiation of the general SMBO framework which uses their new instansification mechanism to iteratively compare its randomly selected parameter configurations against the current incumbent. SMAC can be viewed to be an extension of ROAR which selects configuration based on a model rather than doing it at random.

Another research which described a conceptual framework to optimize hyperparameters is the work by \citet{bergstra2013making}, where they propose a meta modeling approach to support automated hyperparameter tuning. They demonstrated quick recovery of state-of-the art results on several unrelated image classification tasks with no manual intervention. Their framework has four conceptual components: first, a null distribution specification language which describes the configuration distributions; second, a loss function which is the criteria this framework desires to minimize, and which maps a legal configuration sampled form the distribution to a real value; third, the hyperparameter optimization algorithm that takes as input the null distribution and historical values of the loss function and suggests configurations to try next; and fourth, a database that stores experimental history of already-tried configurations.

{\bf{Bayesian Optimization Based Approach:}}
\citet{snoek2012practical} have approached the problem of hyperparameter optimization through Bayesian optimization framework and identified good practices for Bayesian optimization of machine learning algorithms. Bayesian optimization is an iterative algorithm with two key ingredients: a probabilistic surrogate model and an acquisition function to decide which point to evaluate next. \citet{snoek2012practical} argued that a fully Bayesian treatment of the underlying GP kernel is preferred to the approach based on optimization of the GP hyperparameters proposed by previous work. For performing Bayesian optimization, they have selected Gaussian process prior as a prior over the function space and chosen an acquisition function to construct a utility function from model posterior. While describing the algorithm, they have taken into account the availability of multiple processor cores to run experiments in parallel, for which they proposed a sequential strategy. To compute Monte Carlo estimates of the acquisition function under different possible results from pending function evaluations, this strategy takes advantage of the tractable inference properties of GP. 

In a similar manner, \citet{swersky2013multi} have proposed a multi-task Bayesian optimization approach to transfer knowledge within multiple Bayesian optimization frameworks. Applying well-studied multi-task Gaussian process models to Bayesian optimization framework is the basis for this idea. They adaptively learned the degree of correlation among domains by treating a new domain as a new task and then using this information to hone the search algorithm. To perform optimization of related tasks, they restrict their future observations to the tasks of interest; once they have enough observations to estimate the relationship between tasks, then other tasks will act as additional observations without requiring any further functional evaluation. 

{\bf{Gradient Descent Based Approach:}}
\citet{maclaurin2015gradient} have derived a computationally efficient process that will compute hyperparameter gradients using reversed stochastic gradient descent with momentum. They have shown that optimization of validation loss with respect to thousands of hyperparameters can be achieved through these gradients. 

\subsubsection{\textbf{Open Source and Commercial Systems}}

\begin{itemize}[leftmargin=*]
    \item \textbf{Hyperparameter Hunter} \uref{HyperparameterHunter} provides a wrapper for implementing machine learning algorithms by simplifying the experimentation and hyperparameter optimization process. It automatically uses results from past experiments to find optimal hyperparameters. It also eliminates ``boilerplate'' codes and can easily be integrated with other libraries.
    
    \item \textbf{Amazon Sagemaker} \uref{AmazonSagemaker} uses Bayesian Optimization to automatically discover optimal hyperparameters. It starts with a surrogate model and optimizes an special acquisition function called ``Expected Improvement''. In summary, it tries to define and explore a hyperparameter space using Bayesian Optimization and finds optimal hyperparameters that achieves the best model quality. 
    
    \item \textbf{Advisor} \uref{Advisor} is a black box hyperparameter optimization tool. It supports various black box optimization algorithms. Users can choose any algorithm based on their preference.
    
    \item \textbf{Hyperband} \uref{hyperband} uses adaptive resource allocation and early stopping to speedup random search to automatically optimize hyperparameters. Predefined resources are allocated to randomly sampled configurations, and their hyperparameter optimization problem is formulated as a pure-exploration non-stochastic infinite-armed bandit problem.
    
    \item \textbf{BigML OptiML} \uref{BigMLOptiML} uses Bayesian Optimization for hyperparameter tuning which is based on Sequential Model-based Algorithm Configuration (SMAC) optimization technique. It sequentially tries a group of parameters depending on the results of training and evaluating models for those group of parameters. Users can also configure model search optimization metric and select from recommended top performing models.
    
    \item \textbf{Hyperboard} \uref{Hyperboard} facilitates hyperparameter tuning by providing a visualization tool. Users can train models for different set of hyperparameters and plot the performances for each setting. Users can then select one set of hyperparameters by inspecting those plots and selecting the best performing configuration.
    
    \item \textbf{Google HyperTune}  \uref{GoogleHyperTune} uses Bayesian optimization with Gaussian process as the optimization function and ``Expected Improvement'' as their acquisition function. In addition, they try to optimize all currently running jobs.
 
    \item \textbf{SHERPA} \uref{SHERPA} tried to optimize hyperparameters by using Bayesian Optimization via three options: GPyOpt, Asynchronous Successive Halving, and Population-Based Training. Additionally, they support parallel computation according to the user's need, and also provide a live dashboard for results.
 
    \item \textbf{Milano} \uref{Milano} provides automatic hyperparameter optimization and benchmarking  while giving users the option to choose a cloud backend (Azkaban, AWS or SLURM) of their choice. They also allow users to add search algorithms for bench-marking purposes.
    
    \item \textbf{Mind Foundry OPTaaS} \uref{MindFoundryOPTaaS} provides automatic hyperparameter optimization using Bayesian Optimization. They optimize multiple metrics at a time in a secure way (doesn't access user's data or models) and can visualize probabilistic models. 
  
    \item \textbf{BBopt} \uref{BBopt} provides a blackbox optimizer for hyperparameter optimization. They support multiple algorithms for blackbox optimization such as random search, tree structured parzen estimator, Gaussian process, random forest, gradient boosted regression trees, serving, and extra trees. 
   
    \item \textbf{Sigopt} \uref{sigopt} uses an ensemble of Global and Bayesian Optimization algorithms so that they can choose the best optimization technique for the problem given by a user. They choose a suitable algorithm to explore the parameter space under the user's budget constraint. 
   
    \item \textbf{Adatune} \uref{adatune} uses a gradient-based optimizer for hyperparameter tuning. Currently it only supports tuning of the hyperparameter `learning rate,' but can be extended to parameters like `momentum' or `weight decay'. They have used algorithms like `Hypergradient Descent,' `Forward and Reverse Gradient-Based Hyperparameter Optimization,' and `Scheduling the Learning Rate Via Online Hypergradients.'
    
    \item \textbf{Gentun} \uref{gentun} uses genetic algorithms for hyperparameter tuning. They train a model by the set of hyperparameters defined by its genes. They have used a client-server approach to allow multiple clients to train their models, and cross-validation process is handled by a server. Offspring generation and mutation is also managed by the server.  
    
    \item \textbf{Optuna} is an optimization software designed with ``define-by-run'' principle and is particularly the first of its kind. It divides the optimization process into two parts, sampling and pruning. For sampling, they use a blackbox optimization like Bayesian Optimization and grid search. For pruning, they use algorithms like hyperband where they cutoff trials that show little or no promise.
    
    \item \textbf{Test-tube} \uref{test-tube} is a framework agnostic python library that parallelizes hyperparameter search across multiple CPUs and GPUs. They have used grid search and random search as optimization algorithm and user can choose any one of them as they like.

\end{itemize}

\subsection{Automation in Alternative Models Exploration}\label{appen:AME}

\subsubsection{\bf Research Contributions} alternative models exploration is another sub-area where the ML community has invested a lot of effort. These efforts are briefly summarized below.

{\bf{Sequential Model-Based Approach:}}
One school of researchers adopted sequential model-based optimization for automatic selection of learning algorithm and tuning of corresponding hyperparameters. \citet{thornton2013auto} have dubbed this problem as  \emph{Combined Algorithm Selection and Hyperparameter Optimization} (CASH) and approached it with Bayesian optimization, in particular, sequential model-based optimization (SMBO). SMBO first builds a model and uses it to determine candidate configuration of hyperparameters; then, it evaluates the loss and updates the model. They maximize ``Positive Expected Improvement'' as the acquisition function to decide the next useful configuration for evaluation. They combined the popular machine learning framework WEKA with Bayesian optimization to automatically select \textit{good} instantiations of WEKA models, which they named as ``Auto-WEKA''. 

In a similar spirit, \citet{feurer2015efficient} extended the approach of Auto-WEKA \cite{thornton2013auto} in the context of scikit-learn (another popular machine learning tool) and named it ``auto-sklearn''. They take into account the past performances on similar datasets which gives a warm-start to Bayesian Optimization and, from the models already considered by Bayesian optimization, they automatically construct ensembles through an efficient post-processing method. They focused their configuration space on base classifiers while excluding meta-models and ensembles that are themselves parameterized by one or more base classifiers, which resulted in fewer hyperparameters compared to Auto-WEKA.

Another research which used sequential model-based optimization (SMBO) strategy to learn the optimal structure of CNN is the work by \citet{liu2017progressive}. They start with a structured search space and search for convolutional cells instead of a full CNN. A cell consists of multiple blocks, where, a block is a combination operator (such as addition) applied to two inputs (tensors). Depending on training set size and desired runtime of final CNN, the cell structure is stacked. To search the space of cell structure, they proposed a heuristic search which gradually progresses from shallow models to complex models while pruning unpromising structures.

{\bf{Gradient-Based Approach:}}
\citet{zoph2016neural} proposed a gradient based method named ``Neural Architecture Search'' to find good neural-network architectures. They utilized a recurrent network to generate convolutional architectures, and used a controller to generate corresponding architectural hyperparameters. To maximize the expected accuracy of the sampled method, they have trained the neural network with a policy gradient method. They have also used set-selection type attention to enable controller to predict skip connections or branching layers. 

{\bf{Search Space-Based Approach:}}
Many researchers have framed automated model selection as a search problem. \citet{zoph2017learning} defined a search space (NASNet search space) with the property of transferability and searched for the best convolution layer for a small dataset. Then, they applied this cell to a larger data-set by stacking together multiple copies of the cell where each of them will have their own parameters. This way, they designed a convolutional architecture named `NASNet architecture'. To improve generalization in the NASNet models, they have introduced a regularization technique called ``Scheduled Drop Path''. 

Another research group that tried to improve the efficiency of NASNet\cite{zoph2017learning} is \citet{pham2018efficient}, where, they proposed \textit{Efficient Neural Architecture Search} (ENAS) for model discovery. They designed a controller to search for an optimal subgraph within a large computational graph. To select a subgraph that maximizes the expected reward in the validation set, they trained the controller with policy gradient descent that minimized a canonical cross entropy loss. They forced child models to share parameters which allow them to achieve strong empirical performance. 

\citet{liu2017hierarchical} have designed a hierarchical search space where neural network architectures are represented through a hierarchical representation. At each level of the hierarchy, they allowed flexible network topologies. They kept a small set of primitives at the bottom level of the hierarchy and higher-level motifs are formed by lower level motifs. The final neural network is formed by stacking the motifs at the top of the hierarchy. Their work shows that, well designed search space can enhance the results of simple search methods. They have also presented a scalable variant of evolutionary search which improves the results. 

\citet{real2018regularized} used evolutionary algorithm to search the space of NASNet architectures~\cite{zoph2017learning}. They made two additions to the evolutionary search process; first, introducing age property to the tournament selection evolutionary algorithm to favour younger genotypes; second, implementing the set of mutations that NASNet search space would allow to operate. The mutation rules were restricted to only alter the architecture by randomly reconnecting and relabeling edges. They have shown that, specially in the earlier stages of evolution, their approach searched faster than reinforcement learning and random search. 

{\bf{Other Approaches:}}
\citet{lippmann2016d3m} introduced a DARPA program called D3M that aimed to automate model discovery system. They divided the program into two phases. In the first phase, the program built models for empirical science problems with prior complete data where each problem was be supplied with expert generated solutions. In the second phase, the program worked on unspecified and unsolved problems.

\citet{baker2017accelerating} tried to emulate human experts behavior of identifying and terminating suboptimal model configurations through the inspection of partial learning curve. They tried to parameterize learning curve trajectories by extracting simple features from the model architecture, training hyperparameters and early time-series measurements and used these features to train a set of regression models to predict the final validation performance of that configuration. They constructed an early stopping algorithm using these predictions and small model ensembles uncertainty estimates, which helped them to speedup both architecture search and hyperparameter optimization.

\subsubsection{\textbf{Open Source and Commercial Systems}}

\begin{itemize}[leftmargin=*]
    \item \textbf{AutoKeras} \uref{AutoKeras} is an open-source AutoML system for searching better neural architectures with guidance from Bayesian optimization technique. For optimization, they used a neural network as kernel and a tree structure as acquisition function. 

    \item \textbf{AdaNet} \uref{AdaNet} is a tensorflow based framework which require minimal expert intervention. AdaNet provides a framework for meta-learning better models while learning a neural network architecture. It learns the structure of a neural network as well as its weights. 

    \item \textbf{Cloud AutoML} \uref{CloudAutoML} makes machine learning available to users even with limited machine learning knowledge. User can use AutoML to create a custom ML models that are tailored for the users data and integrate those models with the users applications.
    
    \item \textbf{PocketFlow}  \uref{PocketFlow} provides a framework that can compress and accelerate machine learning models. This framework has two components: learners and hyperparameter optimizers. Learners generate a compressed model from the original model with random hyperparameters, and the optimizer evaluates its accuracy and efficiency. After a few iterations, the model with best output is chosen as the final model.
    
    \item \textbf{Automl-gs} \uref{automl-gs} takes an input CSV file, infers the datatype of each attribute, and tries a ETL (extract, transform, and load) strategy for each attribute to create a statistical model. Model ETL and model construction functions are regarded as training scripts and once the model is trained, training results are saved along with hyperparameters. This task is repeated until run count reaches a certain threshold and the best script is saved after each trial.
    
    \item \textbf{MLBox} \uref{MLBox} is an automated machine learning library in python. MLBox can process, clean and format row data in distributed fashion. They use robust methods for feature selection and optimize the hyperparameters in high-dimensional space. For both classical and regression problems, they provide state-of-the art predictive models with interpretation.
    
    \item \textbf{Morph-net} \uref{Morph-net} learns optimal deep network structures during training. They added regularizers which target consumption of specific resources to activate sparsity. They used stochastic gradient descent to minimize summation of regularized loss.
    
    \item \textbf{Determined AI Platform} \uref{DeterminedAIPlatform} helps teams to collaborate by training models with greater speed, sharing GPU resources, and allowing users to build and train models at scale. Their hyperparameter optimizer uses distributed search of the parameter space which helps it quickly find  optimal values for hyperparameters.
    
    \item \textbf{TransmogrifAI} \uref{TransmogrifAI} is built on Scala and SparkML to automate machine learning pipelines. For alternate model selection, TransmogrifAI runs a tournament of different algorithms and automatically chooses the best one by using average validation error. To deal with imbalanced data-sets, it appropriately samples data and recalibrates the prediction . 
    
    \item \textbf{RemixAutoML} \uref{RemixAutoML} can guide users to a baseline starting model quickly. Once  reached to baseline, alternative methods can be explored and compared with the baseline. It utilizes CatBoost, H2O, and XGBoost for machine learning automation jobs.
\end{itemize}

\vspace{-2mm}
\section{\bf Automation in Evaluation}\label{appen:E}
\subsection {\textbf{Research Contributions}}
The machine learning community has spent significant efforts towards  automation of evaluation and benchmarking for ML tasks. \citet{shi2016benchmarking} conducted a benchmarking study of GPU-accelerated deep learning tools. Later, a benchmark study for Reinforcement Learning algorithm's continuous control tasks was reported by \citet{duan2016benchmarking}. In a similar spirit, researchers have introduced publicly available benchmarking dataset suite for supervised classification method (\citet{olson2017pmlb}). In recent studies,  benchmarking of hyperparameter optimization was conducted through surrogates (\citet{eggensperger2015efficient}). In this section, we will briefly explore the research landscape of benchmarking and evaluation for machine learning tasks.

\citet{shi2016benchmarking} aimed at benchmarking GPU-accelerated deep learning tools. They tried to document the running-time performance of some selected deep neural network tools on two CPU platforms and three GPU platforms. Afterwards, they also benchmark some distributed versions on multiple GPUs. From this benchmarking study, they concluded that no single tool consistently outperforms others which indicates further optimization opportunity. 

\citet{duan2016benchmarking} conducted a benchmark study of continuous control tasks for Reinforcement Learning (RL) where they implemented and studied several RL algorithms in the context of general policy parameterizations. Their benchmark consists of 31 continuous control tasks which varies from simple to challenging tasks as well as contains partially observing and hierarchically structured tasks. Although several algorithm like TNPG, TRPO and DDPG were efficient for training deep neural network policies, their poor performance on hierarchical tasks found during the study calls for further improvement.

\citet{olson2017pmlb} introduced a publicly available data-set suite named Penn Machine Learning Benchmark (PMLB) which initially had 165 publicly available data-sets for supervised classification tasks. On the full set of data-sets from PMLB, they conducted performance evaluation of 13 standard machine learning methods from scikit-learn. They have also assessed those benchmarking data-sets diversity based on their predictive performance and from the perspective of their meta features. To generalize ML evaluation, this work integrated real-world, simulated and toy data-sets which make them a pioneer in the assembly of an effective and diverse set of benchmarking standards. 

\citet{eggensperger2015efficient} introduced the interesting idea of automatic benchmarking of hyperparameter optimization through surrogates which share the same hyperparameter space and feature similar response surface. They train regression models on the data of various performances of machine learning algorithm as the hyperparameter settings vary. Then, instead of running the real algorithm, they evaluate hyperparameter optimization methods using the models performance predictions. They have found that tree-based models can capture the performance of multiple machine learning algorithms reasonably well and can yield near real-world benchmarks without running the real algorithms.

\subsection{\textbf{Open Source and Commercial Systems}}

\begin{itemize}[leftmargin=*]
    \item \textbf{OpenML100} \uref{OpenML100} provides in-depth benchmarking of ML experiments on 100 public data-sets and allows users to reproduce the results. They collect experiments created by users and evaluate them with different tools. During evaluation, every algorithm is evaluated with optimized hyperparameter-set and 200 random search iterations.
    
    \item \textbf{OpenML-CC18} \uref{OpenML-CC18} benchmarks on 73 data-sets. For each data-set, they run 1000 random configurations of random forest classifier where training time restricted to maximum three hours.
\end{itemize}

\section{System Automation for ML}\label{appen:system}

\subsection {\textbf{Research Contributions}}
Researchers have built various complex systems that can support data scientists performing different essential sub-tasks within an end-to-end machine learning pipeline. Some researchers used hybrid Bayesian and multi-armed bandit optimization systems (\citet{swearingen2017atm}), where other researchers have leaned towards black-box optimization (\citet{golovin2017google}). Another line of work by \citet{wang2018rafiki} proposes a general programming model which provides a unified system architecture for both training and inference services. In this section, we will briefly review the research article which aim for ML system automation. 

\citet{swearingen2017atm} introduced a machine learning service named Auto-Tuned Models (ATM) that can be hosted on any distributed computing infrastructure. ATM can support multiple users and search through multiple machine learning methods. They have proposed a hierarchical hyperparameter search space containing three levels. The combined hyperparameter search space for a given modeling method is defined by a conditional parameter tree (CPT), and a subset of CPT is defined as a hyperpartition. A hyperpartition contains all choices on the path from root to leaf nodes and a set of tunable conditional hyperparameters is fully defined by the path. \citet{swearingen2017atm} have broken the problem of automatic machine learning process into two sub-problems. First, they select hyperpartition using the bandit learning strategy (specifically, a multi-armed bandit (MAB) framework). Second, they tune the hyperparameters using Gaussian Process based on meta-modeling techniques within a hyperpartition.

\citet{golovin2017google}  introduced `Google Vizier' which is a service for black-box optimization. By default, they use \emph{Batched Gaussian Process Bandits}. They used a Matern kernel with automatic relevance determination. ``Expected improvement'' was chosen as the acquisition function. With a random start, they used a proprietary gradient-free hill climbing algorithm to find local maxima of the acquisition function, and a Gaussian Process regressor was used for uphill walk. Vizier guides and accelerates the current study through prior studies' data by supporting a form of transfer learning.  

\citet{wang2018rafiki} presented a system called `Rafiki' that provides distributed hyperparameter tuning for training service and online ensemble modeling for inference service. They proposed a general programming model to support popular hyperparameter tuning approaches like grid search, random search, and Bayesian optimization. To initialize new trials, they proposed a collaborative tuning scheme which leverages the model parameters of current top-performing training trials. To optimize overall accuracy and reduce latency, they have used reinforcement learning-based scheduling algorithm for the inference service.

\subsection{\textbf{Open Source and Commercial Systems}}

\begin{itemize}[leftmargin=*]
    \item \textbf{DotData} \uref{DotData} provides an automation system for supporting machine learning pipelines. It automatically prepares the given data for feature engineering, explore the feature space, and select relevant features. It derives an accurate predictive model by automatically exploring state-of-the-art algorithms and tuning their hyperparameters.
    
    \item \textbf{ATM} \uref{ATM} is a multi-data system for automated machine learning which was introduced by \citet{swearingen2017atm}.

    \item \textbf{TPOT} \uref{TPOT} is a tree-based optimization tool to automate machine learning pipelines developed by \citet{olson2016evaluation}. TPOT optimizes machine learning pipeline using genetic programming (GP) by maximizing classification accuracy for a given supervised learning data set. TPOT uses GP for evolving the pipeline operator sequences and parameters for maximizing the classification accuracy of the pipeline.
    
    \item \textbf{IBM AutoAI} \uref{IBMAutoAI} provides an automated machine learning model building and evaluation system. AutoAI cleans raw data and categorizes them based on data type. AutoAI tests candidate algorithms against a small subset of the data to rank the algorithms. Gradually, they increase the size of the subset for most promising algorithms. At a high level, AutoAI explores the feature space while maximizing model accuracy using reinforcement learning.
    
    \item \textbf{Auto-Sklearn} \citet{feurer2015efficient} have jointly automated learning algorithm selection and hyperparameter selection and introduced an automated machine learning toolkit called Auto-Sklearn \uref{Auto-Sklearn}. They mainly leveraged Bayesian Optimization, meta-learning and ensemble construction.
    
    \item \textbf{Azure Machine Learning}  \uref{AzureMachineLearning} uses combination of Bayesian optimization and collaborative filtering to solve the meta learning task of machine learning automation. They take uncertainty into account by incorporating a probabilistic model that determines which model to try next.
    
    \item \textbf{Google Cloud AI platform} \uref{GoogleCouldAIplatform}  tests built-in algorithms for datasets submitted by the user. If any built-in algorithm performs well, then that algorithm is selected; otherwise, users can create a training application. This AI Platform Training also uses Bayesian optimization for hyperparameter tuning.
    
    \item \textbf{Amazon AWS service H2O.ai} \uref{AmazoonAWSservice} is Amazon's solution to automate machine learning pipelines. Besides supporting end-to-end automation, it also offers visualization and machine learning interpretability (MLI) functions.
    
    \item \textbf{Data Robot} \uref{DataRobot} finds a good starting point for hyperparameter optimization and users can iterate from there by further customizing learning rate, batch size, network architecture, etc. Data Robot provides state-of-the-art benchmarks to support comparative analysis for the user.
    
    \item \textbf{Amazon Forecast} \uref{AmazonForecast} produces a forecasting model from time series data capable of making future predictions. Their predictions are up to 50\% more accurate than traversing time series data alone.
\end{itemize}

\end{document}